\PassOptionsToPackage{table,dvipsnames}{xcolor}
\documentclass[11pt]{article}

\usepackage[final]{acl}
\usepackage{times}
\usepackage{latexsym}

\usepackage[T1]{fontenc}

\usepackage[utf8]{inputenc}

\usepackage{microtype}

\usepackage{inconsolata}

\usepackage{graphicx}
\usepackage{xcolor}
\usepackage{soul}
\sethlcolor{lightgray}
\usepackage{fontawesome5}

\usepackage{hyperref}
\usepackage{url}
\usepackage{amsmath}   
\usepackage{amssymb}   
\usepackage{amsthm}    
\usepackage{amsfonts}
\usepackage{booktabs}
\usepackage{makecell}
\usepackage{array}
\usepackage{longtable}
\usepackage{latexsym}
\usepackage{booktabs}
\usepackage{adjustbox}
\usepackage{multicol}
\usepackage{multirow}
\usepackage{pdflscape}
\usepackage{philex}
\usepackage{enumitem}
\usepackage{tabularx}
\usepackage{tcolorbox}
\usepackage{subcaption}  
\tcbuselibrary{listings, breakable, skins}

\usepackage{soul}

\definecolor{GoodGreen}{HTML}{2E7D32}
\definecolor{BadRed}{HTML}{C62828}
\definecolor{FinalYellow}{HTML}{FFD966}

\newcommand{\goodstepup}[3]{%
\par\noindent
{\scriptsize\color{GoodGreen!80!black}\textbf{#1}}\\[-0.25em]
{\sethlcolor{GoodGreen!#2!white}\hl{#3}}\par
}

\newcommand{\badstepup}[3]{%
\par\noindent
{\scriptsize\color{BadRed!80!black}\textbf{#1}}\\[-0.25em]
{\sethlcolor{BadRed!#2!white}\hl{#3}}\par
}

\newcommand{\lowesttag}[1]{%
\fcolorbox{#1!70!black}{#1!8!white}{\scriptsize\strut lowest}%
}

\newcommand{\goodloweststepup}[3]{%
\begin{tcolorbox}[
  enhanced,
  breakable,
  boxrule=0.8pt,
  arc=2pt,
  colback=GoodGreen!3!white,
  colframe=GoodGreen!70!black,
  left=4pt,
  right=4pt,
  top=3pt,
  bottom=3pt,
  before skip=0.8em,
  after skip=0.8em
]
{\scriptsize\color{GoodGreen!80!black}\textbf{#1}\hspace{0.5em}\lowesttag{GoodGreen}}\\[-0.25em]
{\sethlcolor{GoodGreen!#2!white}\hl{#3}}
\end{tcolorbox}
}

\newcommand{\badloweststepup}[3]{%
\begin{tcolorbox}[
  enhanced,
  breakable,
  boxrule=0.8pt,
  arc=2pt,
  colback=BadRed!3!white,
  colframe=BadRed!70!black,
  left=4pt,
  right=4pt,
  top=3pt,
  bottom=3pt,
  before skip=0.8em,
  after skip=0.8em
]
{\scriptsize\color{BadRed!80!black}\textbf{#1}\hspace{0.5em}\lowesttag{BadRed}}\\[-0.25em]
{\sethlcolor{BadRed!#2!white}\hl{#3}}
\end{tcolorbox}
}

\newcommand{\finalanswer}[1]{%
\par\noindent
{\sethlcolor{FinalYellow!55!white}\hl{#1}}\par
}

\newtcolorbox{intuitionbox}{
  enhanced,
  breakable,
  colback=yellow!6!white,
  colframe=yellow!35!black,
  coltitle=black,
  fonttitle=\bfseries,
  title={\faLightbulbO\hspace{0.5em}Intuition},
  boxrule=0.4pt,
  arc=3pt,
  left=8pt,
  right=8pt,
  top=6pt,
  bottom=6pt,
  before skip=10pt,
  after skip=10pt
}

\definecolor{pairone}{HTML}{FFF3A3}
\definecolor{pairtwo}{HTML}{B8E1FF}
\definecolor{pairthree}{HTML}{C8F7C5}
\definecolor{pairfour}{HTML}{FFD1DC}

\usepackage[T1]{fontenc}
\usepackage{fix-cm}
\usepackage{times}
\usepackage{amsmath}
\usepackage{tikz}
\usepackage{graphicx}
\usetikzlibrary{calc}

\definecolor{navy}{RGB}{17,35,91}
\definecolor{palePink}{RGB}{255,242,249}
\definecolor{pinkLine}{RGB}{201,85,154}
\definecolor{paleYellow}{RGB}{255,252,230}
\definecolor{yellowLine}{RGB}{221,177,20}
\definecolor{paleBlue}{RGB}{237,247,255}
\definecolor{blueLine}{RGB}{94,168,232}
\definecolor{paleGreen}{RGB}{240,253,241}
\definecolor{greenLine}{RGB}{104,191,116}
\definecolor{paper}{RGB}{253,253,253}

\newcommand{\pagew}{12.62}
\newcommand{\pageh}{8.70}
\newcommand{\colw}{6.05}
\newcommand{\colh}{6.45}
\newcommand{\bodyfont}{\fontsize{7.65}{8.45}\selectfont}
\newcommand{\labelfont}{\bfseries\fontsize{7.85}{8.45}\selectfont}

\newcommand{\tracebox}[3]{%
  \draw[navy,line width=.50pt,rounded corners=2.1pt] (#1,#2) rectangle ++(\colw,\colh);
  \node[text=navy,font=\bfseries\fontsize{11.2}{11.6}\selectfont,anchor=north,inner sep=0pt]
    at (#1+0.5*\colw,#2+6.27) {#3};
  \draw[navy,line width=.50pt] (#1+0.15,#2+5.92) -- (#1+\colw-0.15,#2+5.92);
}

\newcommand{\hilite}[6]{%
  \draw[draw=#5,fill=#4,line width=.32pt,rounded corners=1.8pt] (#1,#2) rectangle ++(#3,#6);
}

\newcommand{\steplineBase}[6]{%
  \node[anchor=north west,text=black,font=\labelfont,inner sep=0pt] at (#1,#2) {#3};
  \node[
    anchor=north west,
    text=black,
    align=left,
    text width=#4,
    font=\bodyfont,
    inner sep=0pt
  ] at (#1+#6,#2) {#5};
}

\newcommand{\stepline}[5]{%
  \steplineBase{#1}{#2}{#3}{#4}{#5}{1.05}
}

\newcommand{\steplineR}[5]{%
  \steplineBase{#1}{#2}{#3}{#4}{#5}{0.9}
}

\title{Boosting Self-Consistency with Ranking}

\author{
 \textbf{Maria Marina\textsuperscript{1,2}},
 \textbf{Daniil Moskovskiy\textsuperscript{1,2}},
 \textbf{Sergey Pletenev\textsuperscript{1,2}},
 \\
 \textbf{Mikhail Salnikov\textsuperscript{1,2}},
 \textbf{Alexander Panchenko\textsuperscript{2,1}},
 \textbf{and Viktor Moskvoretskii\textsuperscript{3}}
\\
 \textsuperscript{1}AIRI,
 \textsuperscript{2}Skoltech,
 \textsuperscript{3}EPFL
\\
\href{mailto:m.marina.scientia@gmail.com}{\texttt{m.marina.scientia@gmail.com}}
}

\begin{document}
\maketitle
\begin{abstract}

Self-consistency improves large language models by sampling multiple reasoning paths and selecting the most frequent answer, but majority voting often fails to recover correct answers that are already present among the samples. We address this limitation with \textit{Ranking-Improved Self-Consistency} (RISC), which reformulates answer selection in self-consistency as a ranking problem. Instead of relying on a single uncertainty or confidence signal, RISC uses a lightweight LambdaRank model to score candidate answers with five carefully designed features that capture answer frequency, semantic centrality, and reasoning-trace consistency. We evaluate RISC on three datasets under a range of test-time budgets. Across datasets, RISC consistently achieves a better accuracy-efficiency trade-off than standard self-consistency and strong baselines, with particularly large gains on question answering benchmarks. Further analysis shows that the proposed features are individually useful and, more importantly, complementary, highlighting the value of learning to combine multiple informative signals for test-time answer selection.

\end{abstract}

\section{Introduction}
Test-time scaling has recently emerged as a promising direction for improving the performance of large language models (LLMs)~\cite{snell2024scalingllmtesttimecompute,muennighoff2025s1simpletesttimescaling}. Instead of relying solely on larger models or more training data, this paradigm improves performance by allocating additional compute during inference. Prior work has shown that additional computation can substantially improve performance through approaches such as extended reasoning~\cite{openai2024openaio1card,Guo_2025}, structured search methods like Tree-of-Thoughts~\cite{yao2023treethoughtsdeliberateproblem}, iterative self-correction~\cite{kamoi-etal-2024-llms,moskvoretskii2025selftaughtselfcorrectionsmalllanguage}, and self-consistency decoding~\cite{self-consistency}.

 \begin{figure}[t!]
    \centering
    \includegraphics[width=\linewidth]{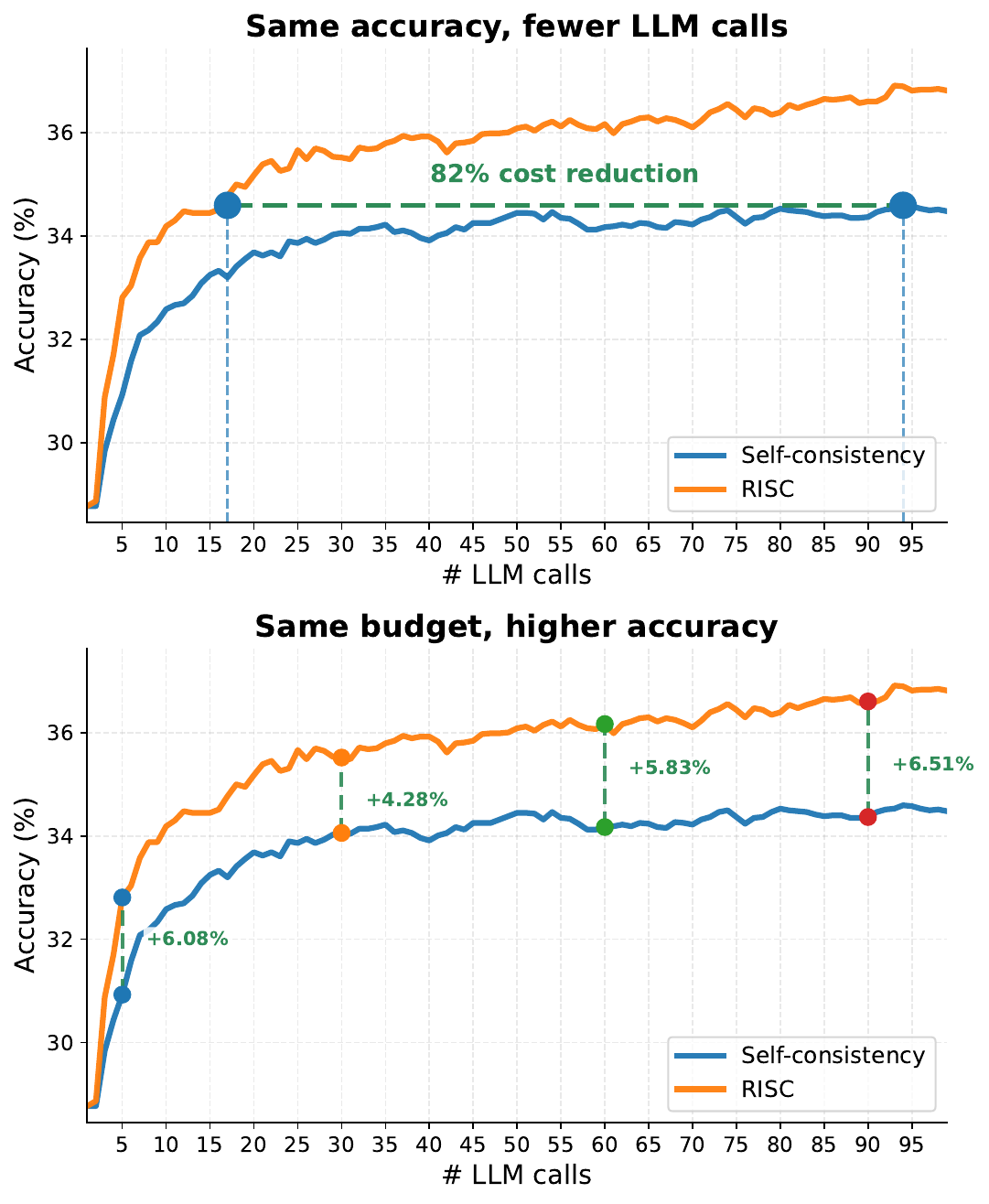}
    \caption{Accuracy versus the number of sampled responses on PopQA for self-consistency and RISC. RISC consistently achieves higher accuracy while substantially reducing computational cost: with only 18 samples, it already surpasses the performance of self-consistency with 99 samples. It also delivers systematic accuracy gains over self-consistency across the full range of LLM-call budgets.
    }
    \label{fig:intro}
    \vspace{-0.8cm}
\end{figure}

In this paper, we focus on improving self-consistency, a test-time scaling technique where an LLM samples multiple candidate answers and selects the final prediction by majority vote~\cite{self-consistency}. Despite its simplicity, self-consistency has proven highly effective while remaining computationally efficient and easy to apply across models, tasks, and hardware~\cite{amballa2025quasi,min2023beyond}.

However, despite its strong performance, a large gap still remains between majority voting and an upper bound where an oracle selects the correct answer from the sampled candidates~\cite{zhuang2026one,dang2025weight}. This gap suggests that the correct answer is often present among the samples, but the model is uncertain and fails to select it reliably.
Several prior works have attempted to identify robust signals that help distinguish the correct answer among sampled candidates. Examples include CISC, which relies on verbal uncertainty estimates~\cite{taubenfeld2025confidence}, and methods based on internal geometric structure, such as stable rank~\cite{tang2025srgrpostablerankintrinsic}.

Nevertheless, these approaches rely on designing a single robust signal intended to capture the model’s parametric knowledge. While appealing in their simplicity, such hand-crafted, knowledge-driven methods often struggle to scale and generalize to new tasks, a pattern long observed in AI research~\cite{sutton2019bitterlesson}.

\begin{figure*}[t!]
    \centering
    \includegraphics[trim=.25cm .35cm .5cm .25cm, clip, width=\linewidth]{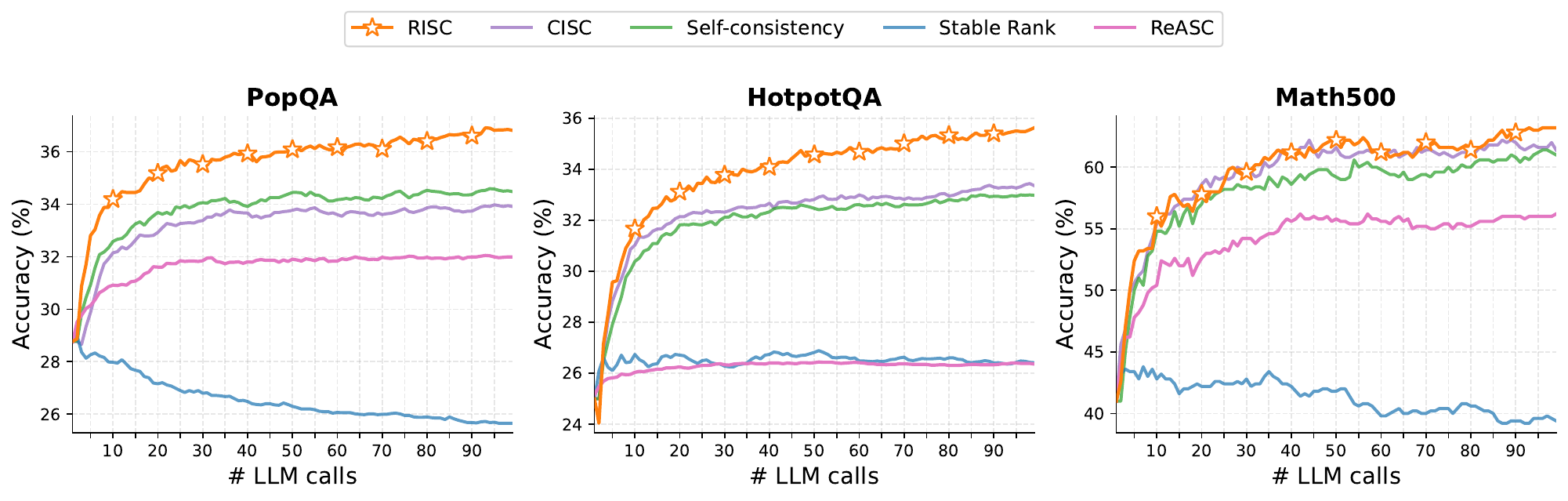}
    \caption{Comparison of RISC against Self-Consistency, Stable Rank, ReASC, and CISC on three datasets for Llama-3.1-8B-Instruct model. RISC consistently outperforms the baselines on the QA datasets across all LLM call budgets, while remaining competitive on MATH500.
    }
    \label{fig:baselines_comparison}
    \vspace{-0.5cm}
\end{figure*}


We therefore introduce \textbf{Ranking-Improved Self-Consistency} (RISC), a lightweight extension of self-consistency. RISC treats answer selection among sampled candidates as a ranking task, where correct answers should appear higher in the output. 


Under this formulation, RISC builds a lightweight ranker that scores candidates using a small set of interpretable features motivated by prior work. This design is aligned with recent QA research showing that lightweight question-level and external-knowledge signals can provide useful reliability information without relying on expensive additional LLM calls~\cite{DBLP:conf/emnlp/MarinaIPSGKKPM25, DBLP:conf/emnlp/PletenevMIGKSKPM25}.

    
    
    


Our results show that RISC significantly outperforms previous methods, while remaining generalizable and performing strongly on challenging test-time scaling tasks such as question answering. We further analyze its behavior and demonstrate that each feature contributes meaningful improvements on its own, while their combination leads to additional performance gains.

The contributions of this paper are as follows:
\begin{enumerate}[leftmargin=8pt, itemsep=0.3pt, topsep=0.2pt]


\item We propose RISC, which reformulates answer selection in self-consistency as a \textit{ranking problem}, and show that RISC consistently outperforms majority voting and strong confidence-based baselines across three benchmarks.

\item We design five interpretable features capturing answer frequency, 
semantic centrality, and reasoning-trace consistency, and show they 
are individually useful and mutually complementary.

\item We provide a detailed analysis via feature ablations and SHAP, 
revealing non-additive interactions between features that explain 
why a learned combination outperforms any single handcrafted signal.

\end{enumerate}

We make code publicly available.\footnote{\url{https://github.com/s-nlp/RISC}}

\section{Related Work}
\paragraph{Test-Time Compute Scaling.}
Recent research has shown that allocating additional computation at inference time can significantly improve the performance of large language models without modifying their parameters~\cite{snell2024scalingllmtesttimecompute}. This family of methods includes parallel strategies that generate multiple candidate outputs and select or combine them.
These techniques have been successful on various tasks across mathematical reasoning~\cite{math500}, code generation~\cite{jimenez2024swebenchlanguagemodelsresolve}, and open-ended question answering~\cite{chen2023universalselfconsistencylargelanguage}. Our work falls within the parallel scaling paradigm: generate $N$ candidate responses and select the best candidate.

\paragraph{Self-Consistency and Majority Voting.}

Self-consistency~\cite{self-consistency} is the most widely used parallel selection strategy: it samples multiple reasoning paths at different temperatures and selects the most frequent final answer by majority vote. It is a strong baseline because diversity across chains can cancel errors. Several extensions have been proposed, including weighted voting based on verbalized or confidence scores and ranked voting methods that use preferences~\cite{chen2023universalselfconsistencylargelanguage, wang2025rankedvotingbasedselfconsistency}. Yet a persistent gap remains between majority-vote accuracy and \textit{oracle selection}~\cite{Wu2024InferenceSL}, the accuracy achievable if one could always pick the correct candidate whenever it is present. This suggests that LLMs often generate correct answers that are not the plurality vote, motivating stronger selection criteria.

\paragraph{Ranking and Learning to Rank.}
Learning-to-rank methods~\cite{burges2006learning,DBLP:journals/ftir/Liu09} have been widely applied in NLP, from passage retrieval to candidate selection in language models. \citet{cobbe2021training} train a verifier to score math solutions, and \citet{math500} extend this with process reward models that evaluate reasoning step by step. In best-of-$N$ selection~\cite{snell2024scalingllmtesttimecompute}, a separately trained scorer ranks $N$ candidates and returns the highest-scoring one.

Our approach is different: we score candidates \emph{relative to the full candidate set} rather than independently, capturing inter-candidate agreement structure; and we use a lightweight gradient-boosted ranker over interpretable features rather than a neural reward model, requiring no additional model beyond the base LLM. In this way, we bring the set-relative perspective of LTR directly into the self-consistency framework, showing that a small set of interpretable features suffices to substantially close the gap to oracle selection.

\section{Ranking-Improved Self-Consistency}

In this section, we present \textit{Ranking-Improved Self-Consistency} (RISC), which casts self-consistency as a ranking problem, and then derive the features that power the ranker.

\subsection{Self-Consistency as a Ranking Problem.}

\paragraph{Task Definition.} Let $x \in \mathcal{X}$ denote an input prompt and $y \in \mathcal{Y}$ a candidate output.
Given a base generative model $p_\theta(y \mid x)$, self-consistency constructs a candidate set by drawing $N$ independent samples,
\begin{equation}
    \mathcal{C}(x) = \{y_1, y_2, \dots, y_N\}, \quad y_i \sim p_\theta(\cdot \mid x).
\end{equation}
The goal is then to select, from $\mathcal{C}(x)$, the candidate that is most likely to yield a correct or otherwise desirable final answer.
We formalize this as a \emph{query-dependent ranking problem}, where the prompt $x$ is the query and the sampled candidates in $\mathcal{C}(x)$ are the items to be ranked.

We seek a scoring function $f_{\phi}$ that assigns each candidate $y \in \mathcal{C}(x)$ 
a score based on the prompt $x$, the candidate itself, and the full candidate set $\mathcal{C}(x)$:
\begin{equation}
    f_{\phi}\bigl(x, y \mid \mathcal{C}(x)\bigr) \in \mathbb{R}.
\end{equation}
The final prediction is then given by
\begin{equation}
    \hat{y} = \arg\max_{y \in \mathcal{C}(x)} f_{\phi}\bigl(x, y \mid \mathcal{C}(x)\bigr).
\end{equation}


In the simplest self-consistency setting, this score is given by the empirical frequency of an answer under repeated sampling, i.e., majority voting~\cite{self-consistency}.
More generally, $f$ may be heuristic or learnable, and may depend on the candidate answer itself, its reasoning trace, or its agreement with other samples.

\paragraph{Learning the Ranker.}
The above-mentioned setting is naturally framed as ranking.
Previous studies found that a reward model may be used as a ranker by scoring candidates independently~\cite{cobbe2021training}, but this is only a special case.
More generally, the objective is to rank candidates within the sampled set $\mathcal{C}(x)$ for a fixed prompt $x$.
This distinction matters because multiple candidates may be valid, semantically equivalent, or lead to the same correct final answer.
Thus, $f(x,y \mid \mathcal{C}(x))$ is better understood as a relative selection score.

Under this formulation, the scoring function is trained to rank candidates within the sampled set, rather than to estimate an absolute reward.
From this perspective, it is often more natural to optimize the ordering of the entire candidate set $\mathcal{C}(x)$ using listwise supervision, rather than relying only on pairwise comparisons.
Given relevance labels or a partial ordering over sampled outputs, a simple listwise objective is
\begin{equation}
\mathcal{L}_{\text{list}} =
- \mathbb{E}_{(x, \mathcal{C}(x))}
\left[
\log
\frac{\exp(f_\phi(x, y^{+}))}
{\sum_{y \in \mathcal{C}(x)} \exp(f_\phi(x, y))}
\right],
\end{equation}
where $y^{+}$ denotes a correct candidate in the sampled set.
More generally, standard listwise learning-to-rank objectives can be used to optimize the ordering of the entire candidate pool.

Overall, this view treats self-consistency as selection from a prompt-specific set of sampled generations.
Its effectiveness depends both on the diversity and quality of the candidate set $\mathcal{C}(x)$ and on how well the scoring function ranks candidates within that set.

\subsection{Training Ranker}

In our study, we train a LightGBM~\cite{lightgbm} LambdaRank~\cite{burges2006learning} model and use NDCG~\cite{10.1145/3130348.3130374} to incorporate list-aware information, placing greater emphasis on errors that more strongly affect the final ordering of the candidate set.

Formally, let $s_i=f_\phi(x,y_i)$ denote the score assigned to candidate $y_i$ in the sampled set $\mathcal{C}(x)$, and let $r_i$ be its relevance label.
Training is based on pairwise comparisons $(y_i,y_j)$ within the same set, but the contribution of each pair is weighted by $|\Delta \mathrm{NDCG}_{ij}|$, the change in the quality of the overall ranked list that would result from swapping their positions. Training details are described in Appendix~\ref{sec:train_details}.

\subsection{Feature descriptions}

For each question, all traces (CoTs) are aggregated at the answer level. In other words, we assess answer quality either directly from the answer itself or indirectly by measuring how cohesive — or conversely, how different—the CoTs associated with the same answer are.

\textbf{Answer length}. Character length of the final answer extracted after CoT.


\hl{\textit{ Intuition} \faLightbulb}

This feature captures whether an answer appears in a clean, compact form. Correct answers are often expressed canonically (for example, as a short number or concise entity name), while incorrect answers are more likely to contain extra formatting, explanatory residue, or malformed text. 

\textbf{Ratio to best}

For question $i$, let,
$n_{i,a}$ be the number of traces producing answer $a$, and let $
N_i = \sum_{a'} n_{i,a'}
$\ be the total number of traces for that question. 

Define the answer share:
$\mathrm{ans\_share_{i,a}} = \frac{n_{i,a}}{N_i}.$

Let the largest answer share for question $i$ be
\[
\mathrm{top1\_share_i}
=
\max_{a'} \mathrm{ans\_share_{i,a'}}
\]
Then, ratio-to-best is 
\vspace{-0.1cm}
\[
\mathrm{ratio\_to\_best_{i,a}}
=
\frac{\mathrm{ans\_share_{i,a}}}{\mathrm{top1\_share_i}}.
\]

\hl{\textit{ Intuition} \faLightbulb}

This feature measures how competitive an answer is relative to the strongest answer for the same question. It helps the ranker understand whether a candidate is nearly tied with the dominant answer or far behind it. 

\textbf{Distance from answer to answer centroid}

Let  $V_{i,a} \in \mathbb{R}^d $ denote the embedding of candidate answer $a$ for question $i$. 

We define the count-weighted centroid for question $i$ as 
\vspace{-0.1cm}

\[
C_i^{\mathrm{ans}}
=
\frac{\sum_{a'} n_{i,a'} V_{i,a'}}{\sum_{a'} n_{i,a'}}.
\]

The feature is the squared Euclidean distance from the answer embedding to this weighted centroid:

$\mathrm{ans\_centroid\_dist_{i,a}}
=
\left\| V_{i,a} - C_i^{\mathrm{ans}} \right\|_2^2.
$

\hl{\textit{Intuition} \faLightbulb}

This feature measures how far a candidate answer lies from the semantic center of all answers proposed for the same question. Weighting the centroid by answer frequency makes it a “center of mass” of the answer distribution rather than an unweighted average over unique strings. Answers that are semantically central to the candidate set may be more plausible, while scattered or atypical answers are often less reliable.

\begin{figure}[htb!]
    \centering
    \includegraphics[trim=1.25cm 0.5cm 2.2cm 0cm, width=0.9\linewidth]{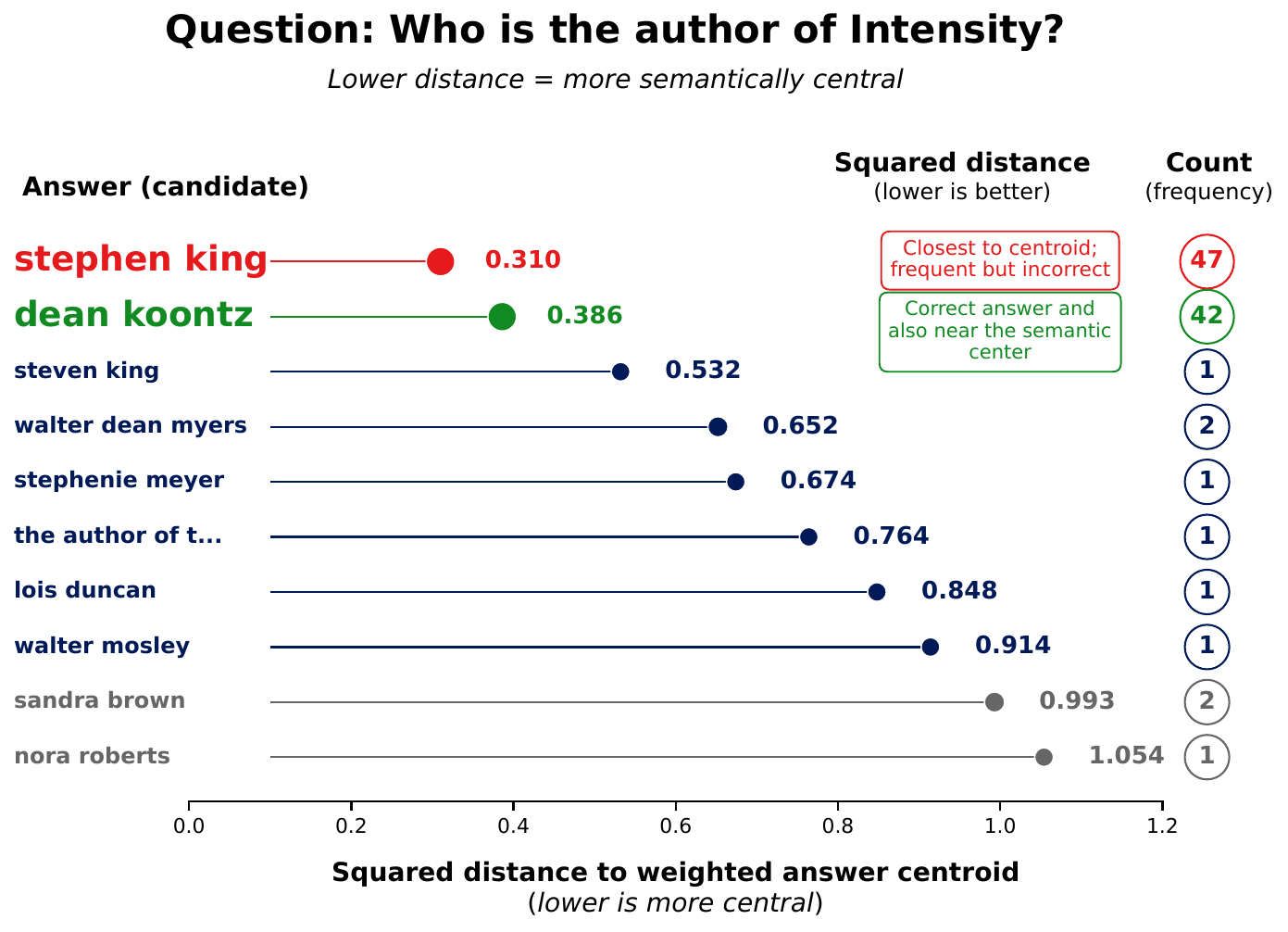}
    \caption{
    Example visualization of the answer-centroid distance feature. Each point shows the squared distance from a candidate answer embedding to the count-weighted answer centroid; lower values indicate greater semantic centrality.}
    \label{fig:ans-centroid-dist}
\end{figure}

\textbf{Worst-step coherence}

For a trace \(t\), let its step embeddings be $
s_{t,1}, s_{t,2}, \dots, s_{t,m_t} \in \mathbb{R}^d .$ Define the within-trace step centroid as $
c_t=\frac{1}{m_t}\sum_{j=1}^{m_t} s_{t,j}.$

For each step, compute its cosine similarity to the trace centroid: $
r_{t,j}=\cos(s_{t,j}, c_t).$ The trace-level worst-step coherence is $
r_t^\star=\min_{1 \le j \le m_t} r_{t,j}.$

For a fixed question \(i\) and answer \(a\), let \(T_{i,a}\) be the set of traces producing answer \(a\). The answer-level feature is the mean of the trace-level worst-step coherences: 

\[\mathrm{worst\_step\_coherence_{i,a}} =
\frac{1}{|T_{i,a}|} \sum_{t \in T_{i,a}} r_t^\star.\]

\hl{\textit{ Intuition} \faLightbulb}


This feature captures whether every step in a reasoning trace stays semantically aligned with the overall trajectory. A low value suggests that at least one step is off-direction, which may indicate a derailment even if the final answer looks plausible.

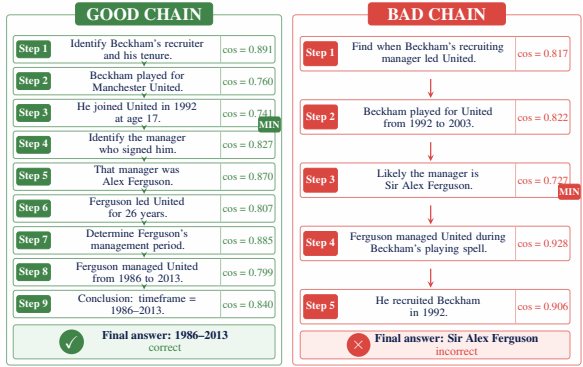
\begin{figure}[htb!]
    \centering
    \begin{adjustbox}{
        trim={2cm .2cm 0.25cm .2cm},
        clip,
        width=\columnwidth,
        keepaspectratio
    }
        \definecolor{gbnavy}{RGB}{15,31,76}
\definecolor{gbpaper}{RGB}{254,254,254}
\definecolor{gbgood}{RGB}{64,132,73}
\definecolor{gbgoodBorder}{RGB}{124,174,132}
\definecolor{gbgoodSoft}{RGB}{238,247,239}
\definecolor{gbbad}{RGB}{218,70,64}
\definecolor{gbbadBorder}{RGB}{232,128,124}
\definecolor{gbbadSoft}{RGB}{253,237,235}
\definecolor{gbrowfill}{RGB}{254,254,254}

\def\gbpagew{13.30}
\def\gbpageh{9.75}
\def\gbmainfont{\fontsize{7.2}{7.75}\selectfont}
\def\gbcosfont{\fontsize{7.0}{7.5}\selectfont}
\def\gbstepfont{\bfseries\fontsize{7.0}{7.2}\selectfont}

\newcommand{\gbchainpanel}[5]{%
  \draw[draw=#5,line width=.42pt,rounded corners=2.1pt] (#1,#2) rectangle ++(#3,#4);
}

\newcommand{\gbchainlabel}[6]{%
  \draw[draw=#5,fill=#5,line width=.35pt,rounded corners=2pt] (#1,#2) rectangle ++(#3,#4);
  \node[text=white,font=\bfseries\fontsize{11.2}{11.6}\selectfont,anchor=center,inner sep=0pt]
    at (#1+0.5*#3,#2+0.5*#4) {#6};
}

\newcommand{\gbchainrow}[9]{%
  \pgfmathsetmacro{\badgecx}{#1+0.46}%
  \pgfmathsetmacro{\midy}{#2+0.5*#4}%
  \pgfmathsetmacro{\textcx}{#1+0.93+0.5*(#3-2.18)}%
  \pgfmathsetmacro{\cosx}{#1+#3-0.61}%
  \pgfmathsetmacro{\vlinex}{#1+#3-1.23}%
  \draw[draw=#5!70!white,fill=gbrowfill,line width=.36pt,rounded corners=1.5pt]
    (#1,#2) rectangle ++(#3,#4);
  \draw[draw=#5!45!white,line width=.30pt] (\vlinex,#2) -- ++(0,#4);
  \draw[draw=#5,fill=#5,line width=.30pt,rounded corners=1.5pt]
    (#1+0.08,#2+0.12) rectangle ++(0.76,#4-0.24);
  \node[text=white,font=\gbstepfont,anchor=center,inner sep=0pt]
    at (\badgecx,\midy) {Step #6};
  \node[text=gbnavy,font=\gbmainfont,align=center,text width=#9,anchor=center,inner sep=0pt]
    at (\textcx,\midy) {#7};
  \node[text=#5,font=\gbcosfont,anchor=center,inner sep=0pt]
    at (\cosx,\midy) {cos = #8};
}

\newcommand{\gbgoodrow}[7]{%
  \gbchainrow{#1}{#2}{#3}{#4}{gbgood}{#5}{#6}{#7}{3.35cm}%
}

\newcommand{\gbbadrow}[7]{%
  \gbchainrow{#1}{#2}{#3}{#4}{gbbad}{#5}{#6}{#7}{3.90cm}%
}

\newcommand{\gbfinalrow}[9]{%
  \pgfmathsetmacro{\iconx}{#1+1.32}%
  \pgfmathsetmacro{\midy}{#2+0.5*#4}%
  \pgfmathsetmacro{\textx}{#1+0.58*#3}%
  \pgfmathsetmacro{\texty}{#2+0.52*#4}%
  \draw[draw=#6,fill=#5,line width=.40pt,rounded corners=1.8pt]
    (#1,#2) rectangle ++(#3,#4);
  \node[circle,draw=none,fill=#7,text=white,font=\bfseries\fontsize{11}{11}\selectfont,inner sep=1.4pt]
    at (\iconx,\midy) {$#8$};
  \node[text=gbnavy,font=\fontsize{7.6}{8.1}\selectfont,align=center,anchor=center,inner sep=0pt]
    at (\textx,\texty) {#9};
}

\newcommand{\gbmintag}[4]{%
  \draw[draw=#3,fill=#3,line width=.30pt,rounded corners=1.4pt] (#1,#2) rectangle ++(0.49,0.34);
  \node[text=white,font=\bfseries\fontsize{6.4}{6.6}\selectfont,inner sep=0pt]
    at (#1+0.245,#2+0.17) {#4};
}

\begin{tikzpicture}[x=1cm,y=1cm]
\path[use as bounding box] (0,0) rectangle (\gbpagew,\gbpageh);
\fill[gbpaper] (0,0) rectangle (\gbpagew,\gbpageh);

\node[
  text=gbnavy,
  font=\bfseries\fontsize{12.7}{13.2}\selectfont,
  align=center,
  anchor=north,
  inner sep=0pt
]
  at (0.5*\gbpagew,9.55)
  {Question: The football manager who recruited David Beckham\\managed Manchester United during what timeframe?};

\gbchainpanel{0.15}{0.23}{6.22}{7.92}{gbgoodBorder}
\gbchainpanel{6.62}{0.23}{6.52}{7.92}{gbbadBorder}

\gbchainlabel{1.71}{7.92}{3.10}{.5}{gbgood}{GOOD CHAIN}
\gbchainlabel{8.33}{7.92}{3.10}{.5}{gbbad}{BAD CHAIN}

\gbgoodrow{0.30}{7.06}{5.92}{0.62}{1}{Identify Beckham's recruiter\\and his tenure.}{0.891}
\gbgoodrow{0.30}{6.34}{5.92}{0.62}{2}{Beckham played for\\Manchester United.}{0.760}
\gbgoodrow{0.30}{5.62}{5.92}{0.62}{3}{He joined United in 1992\\at age 17.}{0.741}
\gbgoodrow{0.30}{4.90}{5.92}{0.62}{4}{Identify the manager\\who signed him.}{0.827}
\gbgoodrow{0.30}{4.18}{5.92}{0.62}{5}{That manager was\\Alex Ferguson.}{0.870}
\gbgoodrow{0.30}{3.46}{5.92}{0.62}{6}{Ferguson led United\\for 26 years.}{0.807}
\gbgoodrow{0.30}{2.74}{5.92}{0.62}{7}{Determine Ferguson's\\management period.}{0.885}
\gbgoodrow{0.30}{2.02}{5.92}{0.62}{8}{Ferguson managed United\\from 1986 to 2013.}{0.799}
\gbgoodrow{0.30}{1.30}{5.92}{0.62}{9}{Conclusion: timeframe =\\1986--2013.}{0.840}

\gbfinalrow{0.30}{0.37}{5.92}{0.77}
  {gbgoodSoft}{gbgoodBorder}{gbgood}{\checkmark}
  {\textbf{Final answer: 1986--2013}\\{\color{gbgood}correct}}

\foreach \ys/\ye in {
  7.04/6.98,
  6.32/6.26,
  5.60/5.54,
  4.88/4.82,
  4.16/4.10,
  3.44/3.38,
  2.72/2.66,
  2.00/1.94
}{%
  \draw[gbgood,->,line width=.65pt,>=stealth] (3.26,\ys) -- (3.26,\ye);
}

\gbmintag{5.85}{5.5}{gbgood}{MIN}

\gbbadrow{6.79}{6.88}{6.10}{0.78}{1}{Find when Beckham's recruiting\\manager led United.}{0.817}
\gbbadrow{6.79}{5.45}{6.10}{0.78}{2}{Beckham played for United\\from 1992 to 2003.}{0.822}
\gbbadrow{6.79}{4.02}{6.10}{0.78}{3}{Likely the manager is\\Sir Alex Ferguson.}{0.727}
\gbbadrow{6.79}{2.59}{6.10}{0.78}{4}{Ferguson managed United during\\Beckham's playing spell.}{0.928}
\gbbadrow{6.79}{1.16}{6.10}{0.78}{5}{He recruited Beckham\\in 1992.}{0.906}

\gbfinalrow{6.79}{0.37}{6.10}{0.63}
  {gbbadSoft}{gbbadBorder}{gbbad}{\times}
  {\textbf{Final answer: Sir Alex Ferguson}\\{\color{gbbad}incorrect}}

\draw[gbbad,->,line width=.80pt,>=stealth] (9.74,6.70) -- (9.74,6.27);
\draw[gbbad,->,line width=.80pt,>=stealth] (9.74,5.27) -- (9.74,4.84);
\draw[gbbad,->,line width=.80pt,>=stealth] (9.74,3.84) -- (9.74,3.41);
\draw[gbbad,->,line width=.80pt,>=stealth] (9.74,2.41) -- (9.74,1.98);

\gbmintag{12.63}{3.98}{gbbad}{MIN}

\end{tikzpicture}
    \end{adjustbox}
    \caption{
    Example visualization of the worst-step coherence feature. Each row shows a reasoning step with its cosine similarity to the within-chain step centroid; lower values indicate weaker alignment with the chain’s overall reasoning trajectory. The minimum-coherence step is marked with MIN, and final answer boxes show whether the chain produced the correct answer. Step sentences are shortened with ellipses for readability. The full example is provided in Appendix~\ref{appendix:worst_step}.
    }
    \label{fig:worst-step-coherence}
\end{figure}

\textbf{Shared checkpoints count}

For a fixed question $i$ and answer $a$, consider the set of traces $T_{i,a}$. Each trace is converted into a sequence of prefixes. If trace $t$ has steps $x_{t,1}, x_{t,2}, \dots, x_{t,m_t},$ then its prefixes are 
\[
\begin{aligned}
p_{t,1}   &= x_{t,1},\\
p_{t,2}   &= x_{t,1} \oplus x_{t,2}, \cdots\\
p_{t,m_t} &= x_{t,1} \oplus x_{t,2} \oplus \cdots \oplus x_{t,m_t}.
\end{aligned}
\]

Let $e_{i,j}$ be the embedding of prefix $p_{t,j}$. Let the normalised depth of prefix $j$ in trace $t$ be $d_{t,j} = \frac{j}{m_t}$

Two prefixes $p_{t,j}$ and $p_{t',k}$ between traces are considered to be matched if they share two conditions:

\begin{enumerate}
    \item |$d_{t,j} - d_{t',k}| \le \tau_{\mathrm{depth}}$
    \item $
\cos(e_{t,j}, e_{t',k}) \ge \tau_{\mathrm{sim}}$

\end{enumerate}
A prefix $p_{t,j}$
 is called a shared checkpoint, denoted $z_{t,j}$, if it matches prefixes from at least a fraction $q$ of the other traces with the same final answer:
\vspace{-0.1cm}

\[
\mathrm{shared\_checkpoints_{i,a}}
=
\sum_{t \in T_{i,a}} \sum_{j=1}^{m_t} z_{t,j}.
\]

\hl{\textit{Intuition} \faLightbulb}


This feature measures how much intermediate reasoning structure is shared among traces with the same answer. It asks not only whether traces reach the same final answer, but also whether they pass through similar semantic states at similar stages. Higher values indicate a more stable and reproducible reasoning pattern.


\begin{figure}[htb!]
    \centering
    \begin{adjustbox}{
        trim={0cm .35cm .2cm .3cm},
        clip,
        max width=\columnwidth, 
        max height=\textheight, 
        keepaspectratio}
        \begin{tikzpicture}[x=1cm,y=1cm]
\path[use as bounding box] (0,0) rectangle (\pagew,\pageh);
\fill[paper] (0,0) rectangle (\pagew,\pageh);

\node[text=navy,font=\bfseries\fontsize{19.6}{19.8}\selectfont,anchor=north,inner sep=0pt]
  at (0.5*\pagew,8.50) {Factoring $x^8 + 3x^4 - 4$};

\node[text=black,font=\itshape\fontsize{9.0}{9.4}\selectfont,anchor=north,inner sep=0pt]
  at (0.5*\pagew,7.85) {Two different solution traces with matched checkpoints.};

\draw[navy,line width=.50pt,rounded corners=2.1pt] (0.70,6.86) rectangle (11.98,7.44);

\node[text=navy,font=\bfseries\fontsize{9.7}{10.2}\selectfont,anchor=west,inner sep=0pt]
  at (1.48,7.15) {Problem:};

\node[text=black,font=\fontsize{9.2}{9.4}\selectfont,anchor=west,inner sep=0pt]
  at (2.94,7.15) {Factor the polynomial $x^8 + 3x^4 - 4$ completely over the integers.};

\tracebox{0.15}{0.35}{Trace A}
\tracebox{6.42}{0.35}{Trace B}

\hilite{0.30}{5.18}{5.76}{palePink}{pinkLine}{0.82}
\hilite{0.30}{2.64}{5.76}{paleYellow}{yellowLine}{0.86}
\hilite{0.30}{1.67}{5.76}{paleBlue}{blueLine}{0.82}
\hilite{0.30}{0.64}{5.76}{paleGreen}{greenLine}{0.72}

\stepline{0.41}{5.82}{Step 1:}{4.58cm}{Factor the polynomial and then\\evaluate ...}
\stepline{0.41}{4.90}{Step 2:}{4.68cm}{Note $x^8\!+\!3x^4\!-\!4\!=\!(x^4)^2\!+\!3x^4\!-\!4$.}
\stepline{0.41}{4.18}{Step 3:}{4.58cm}{Factor the quadratic in $x^4$:\\$(x^4\!+\!4)(x^4\!-\!1).$}
\stepline{0.41}{3.32}{Step 4:}{4.58cm}{Obtain the factorization\\$(x^2+x+1)(x^2-x+1)\ldots$}
\stepline{0.41}{2.34}{Step 5:}{4.58cm}{Evaluate the factors at $x=1$,\\giving $3,\;1$ and $5$.}
\stepline{0.41}{1.17}{Step 6:}{4.58cm}{Sum the values to get $9$.}

\hilite{6.57}{5.18}{5.76}{palePink}{pinkLine}{0.82}
\hilite{6.57}{2.05}{5.76}{paleYellow}{yellowLine}{1.02}
\hilite{6.57}{1.16}{5.76}{paleBlue}{blueLine}{0.82}
\hilite{6.57}{0.4}{5.76}{paleGreen}{greenLine}{0.78}

\steplineR{6.68}{5.82}{Step 1:}{4.95cm}
  {Use a factorization strategy\\to factor completely ...}

\steplineR{6.68}{4.90}{Step 2:}{4.95cm}
  {Write $x^8\!+\!3x^4\!-\!4=(x^4\!+\!4)(x^4\!-\!1)$.}

\steplineR{6.68}{4.5}{Step 3:}{4.95cm}
  {Factor $x^4\!-\!1$ as\\
   $(x^2\!+\!1)(x\!+\!1)(x\!-\!1)$.}

\steplineR{6.68}{3.8}{Step 4:}{4.95cm}
  {Factor $x^4\!+\!4$ as\\
   $(x^2\!-\!2x\!+\!2)(x^2\!+\!2x\!+\!2)$.}

\steplineR{6.68}{2.95}{Step 5:}{4.95cm}
  {List the irreducible factors:\\
   $x^2\!-\!2x\!+\!2,\;x^2\!+\!2x\!+\!2,\;x^2\!+\!1,$\\
   $x\!+\!1,\;x\!-\!1$.}

\steplineR{6.68}{1.79}{Step 6:}{4.95cm}
  {Substitute $x=1$ into each factor\\
   to obtain $1,\;5,\;2,\;2,$ and $0$.}

\steplineR{6.68}{0.9}{Step 7:}{4.95cm}
  {Then $\sum_{i=1}^{5}p_i(1)=10$.}

\end{tikzpicture}
    \end{adjustbox}
    \caption{Example visualization of the shared-checkpoints feature. Matching colors indicate semantically shared checkpoints detected at similar reasoning depths across two reasoning traces. Step sentences are shortened with ellipses for readability. Full example is provided in Appendix~\ref{appendix:shared_checkpoints}.}
    \label{fig:shared-checkpoints}
\end{figure}
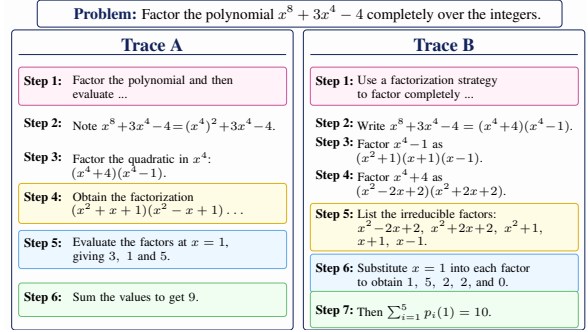

We encode answers, chain-of-thoughts, prefixes, and reasoning steps using the MiniLM\footnote{\url{hf.co/sentence-transformers/all-MiniLM-L6-v2}} sentence embedding model~\cite{minilm} (just 22.7M parameters), which produces dense sentence-level embeddings via transformer encoding followed by pooling.

\section{Experimental Setup}

\subsection{Datasets}

In this work, we utilize a suite of datasets to cover complex tasks: long-tail QA, multi-hop reasoning and math reasoning. \textbf{PopQA}~\cite{popqa} is an open-domain question answering benchmark consisting of entity-centric questions derived from Wikipedia. Each question is associated with a popularity score reflecting the monthly Wikipedia page view count of the corresponding entity. Following~\citet{popqa}, we partition the dataset in train and test set. \textbf{HotpotQA}~\cite{hotpotqa} requires multi-hop reasoning over multiple supporting documents. In our work, we use the closed-book scenario, i.e., without gold or distractor contexts. \textbf{MATH500}~\cite{math500} is a subset of 500 competition-level mathematics problems spanning algebra, geometry, number theory, and other domains. 





\subsection{Baselines}

\paragraph{Self-Consistency.}
Self-Consistency~\cite{self-consistency} samples $m$ reasoning chains and selects the answer by majority vote:
\begin{equation}
    \hat{a}_{\mathrm{SC}} = \arg\max_{a} \sum_{i=1}^{m} \mathbf{1}[a_i = a].
\end{equation}
It requires no additional training or auxiliary models, but achieving reliable performance demands a large number of samples.

\paragraph{Confidence-Informed Self-Consistency.}

CISC~\cite{taubenfeld2025confidence} replaces uniform votes with confidence-weighted aggregation. A per-response confidence score $c_i$ is obtained via the $\mathbb{P}(\text{True})$ method~\cite{kadavath2022language} and normalized via temperature-scaled softmax, yielding:
\begin{equation}
    \hat{a}_{\mathrm{CISC}} = \arg\max_{a} \sum_{i=1}^{m} \mathbf{1}[a_i = a] \cdot \tilde{c}_i.
\end{equation}
CISC reduces the required number of samples by over 40\% on average while matching SC accuracy.

\paragraph{Stable Rank.}
\citet{tang2025srgrpostablerankintrinsic} propose using the stable rank of the response hidden-state matrix $\mathbf{H} \in \mathbb{R}^{T \times d}$ as a reference-free quality signal:
\begin{equation}
    \mathrm{SR}(\mathbf{H}) = \frac{\|\mathbf{H}\|_F^2}{\|\mathbf{H}\|_2^2},
\end{equation}
observing that higher stable rank correlates with response quality in best-of-$N$ selection.


\paragraph{ReASC.}
ReASC~\cite{reasc} replaces count-based stopping with a reliability-aware evidence sufficiency criterion. It first attempts to answer from a single sample, accepting the response only if its confidence exceeds a calibrated threshold. Otherwise, it continues sampling and updates a Beta posterior with confidence-weighted pseudo-counts, so higher-confidence responses contribute more evidence than lower-confidence ones. Sampling stops when the posterior indicates sufficient support for the leading answer. This lets ReASC avoid unnecessary samples and reduce inference cost.

\begin{table*}[ht!]
\centering
\footnotesize
\begin{tabular}{lccccccccc}
\toprule
& \multicolumn{3}{c}{\textbf{Cost Reduction}}
& \multicolumn{3}{c}{\textbf{Accuracy Improvement}}
& \multicolumn{3}{c}{\textbf{Headroom Over Maximum}} \\
\cmidrule(lr){2-4} \cmidrule(lr){5-7} \cmidrule(lr){8-10}
\textbf{Dataset}
& \textbf{@5} & \textbf{@10} & \textbf{@15}
& \textbf{@5} & \textbf{@50} & \textbf{@99}
& \textbf{@50} & \textbf{@75} & \textbf{@99} \\
\midrule
PopQA    & 61.54 & 71.43 & 70.00 & 6.08 & 4.75 & 6.77 & 4.31 & 5.34 & 6.42 \\
HotpotQA & 37.50 & 50.00 & 64.29 & 5.95 & 6.49 & 8.03 & 4.79 & 6.63 & 7.98 \\
Math500  & 37.50 & 28.57 & 28.57 & 4.80 & 5.07 & 3.61 & 1.30 & 0.33 & 2.93 \\
\midrule
Average  & 45.51 & 50.00 & 54.29 & 5.61 & 5.44 & 6.14 & 3.47 & 4.10 & 5.78 \\
\bottomrule
\end{tabular}
\caption{Performance summary on PopQA, HotpotQA, and Math500 under different LLM-call budgets. Here, @\(k\) denotes results at a budget of \(k\) LLM calls. We report cost reduction, accuracy improvement, and headroom over the maximum, where headroom is defined relative to SC’s maximum achieved accuracy.}
\label{tab:cost_accuracy_headroom}
\end{table*}

 \begin{figure*}[t!]
    \centering
    \includegraphics[width=\linewidth]{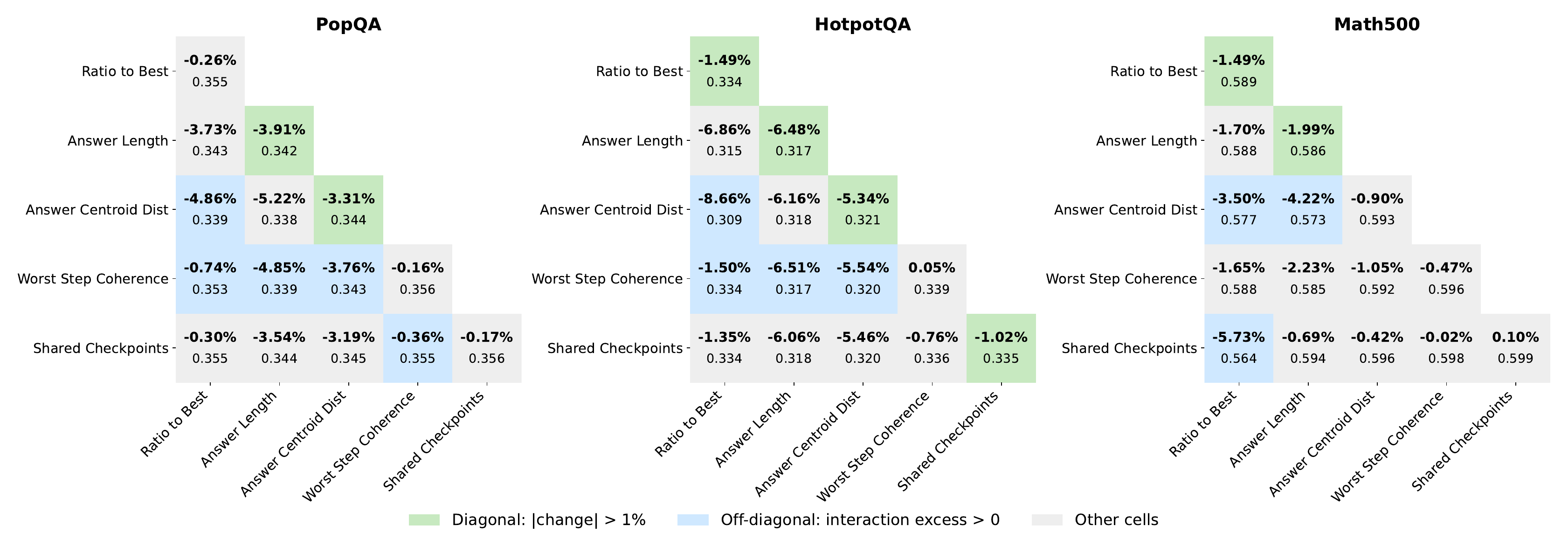}
    \caption{\textbf{Feature ablation across three datasets.} Heatmaps show the percentage change in mean accuracy, averaged over budgets 1–99 LLM calls, relative to the full ranker. Diagonal cells represent single-feature ablations, and lower-triangular cells represent two-feature ablations. Raw mean accuracies for the ablated models are shown in parentheses. Cells are colored by effect type: diagonal cells with an absolute drop greater than 1\% are highlighted, while off-diagonal cells are highlighted when the joint ablation is larger than the sum of individual drops. All other cells are shown in light gray.}
    \label{fig:ablation}
    \vspace{-0.5cm}
\end{figure*}

\subsection{Experimental Details}

\paragraph{Generation Setup.} In all our experiments we use Llama 3.1 8B Instruct\footnote{\url{hf.co/meta-llama/Llama-3.1-8B-Instruct}}~\cite{llama3}. As an additional generalizability check, we evaluate our method on the Olmo-3-7B-Instruct model; the results are reported in Appendix~\ref{fig:baselines_comparison_olmo}\footnote{\url{hf.co/allenai/Olmo-3-7B-Instruct}}~\cite{olmo2025olmo3}. This model shows extensive knowledge coverage and benefits most from scaling on all datasets. For all datasets, samples were generated with a temperature of t=0.9, top\_p=1.0, we generated 99 samples sequentially. The generation for the train and test splits was done at 0-shot with Chain-of-Thought~\cite{cot}; prompts are available in Appendix~\ref{sec:prompts}. 

\paragraph{Evaluation Setup.} We evaluate QA performance using Inclusion Accuracy~(In-Acc)~\cite{drqa} as a relaxed exact match by checking  whether any reference answer occurs as an adjacent token subsequence  within the extracted final answer $\hat{a}$. 




\paragraph{Efficiency and Accuracy Metrics.}
Following CISC~\cite{taubenfeld2025confidence}, we additionally evaluate response selection methods using two complementary metrics. 
The \emph{cost reduction} is defined as:
\vspace{-0.5cm}
\begin{equation}
    \text{Cost Reduction} = 100 \times \left(1 - \frac{b}{b_{\mathrm{SC}}}\right)\%,
\end{equation}
where $b$ is a fixed sample budget for a method and $b_{\mathrm{SC}}$ is the number of calls that standard self-consistency requires to reach the same accuracy level. 
The second metric captures the accuracy gain when all methods are given the same number of responses. Given accuracy $\text{Acc}_{\mathrm{method}}$ and $\text{Acc}_{\mathrm{SC}}$ at equal sample budgets, the \emph{accuracy improvement} is:
\vspace{-0.2cm}
\begin{equation}
    \text{Acc.\ Improvement} = 100 \times \left(\frac{\text{Acc}_{\mathrm{method}}}{\text{Acc}_{\mathrm{SC}}} - 1\right)\%.
\end{equation}
For the QA datasets, we observe that beyond a certain budget RISC attains an accuracy that self-consistency never matches, even at the maximum number of sampled calls. To quantify this gap, we introduce a third metric, the \emph{headroom over maximum SC}, defined as:
\begin{equation}
    \ 100 \times \left(
    \frac{\text{Acc}_{\mathrm{method}}(b)}
    {\max_{k \leq K} \text{Acc}_{\mathrm{SC}}(k)}
    - 1
    \right)\%.
\end{equation}

Together, these metrics allow us to compare methods both in terms of how much computation they save and how much accuracy they add at fixed cost. 

\section{Results}

Figure~\ref{fig:baselines_comparison} shows a consistent pattern across all three datasets: RISC improves steadily with budget and stays above both CISC and vanilla self-consistency over most of the range, while Stable Rank is clearly weaker and sometimes degrades as more calls are added. On PopQA and HotpotQA, the gap appears early and widens with budget. On MATH500, the gap is smaller, but RISC remains competitive at low budgets and finishes highest at the largest budgets. The advantage is therefore not limited to low budgets: across all three datasets, RISC remains strong or continues improving at high budgets, rather than saturating as early as CISC or self-consistency.

The quantitative results in Table~\ref{tab:cost_accuracy_headroom} support this pattern: RISC consistently provides a better efficiency–accuracy trade-off than the baselines across all three datasets. Positive cost-reduction values show that it often matches self-consistency with substantially fewer calls, especially on PopQA and HotpotQA. The accuracy-improvement metric is positive in every reported setting, showing that the method is not only cheaper at matched quality but also more accurate at the same budget. Finally, positive headroom-over-maximum values show that in many settings RISC reaches accuracy levels that self-consistency never attains within the sampled range, particularly on the QA datasets.

\begin{figure}[t!]
    \centering
    \includegraphics[width=0.65\columnwidth]{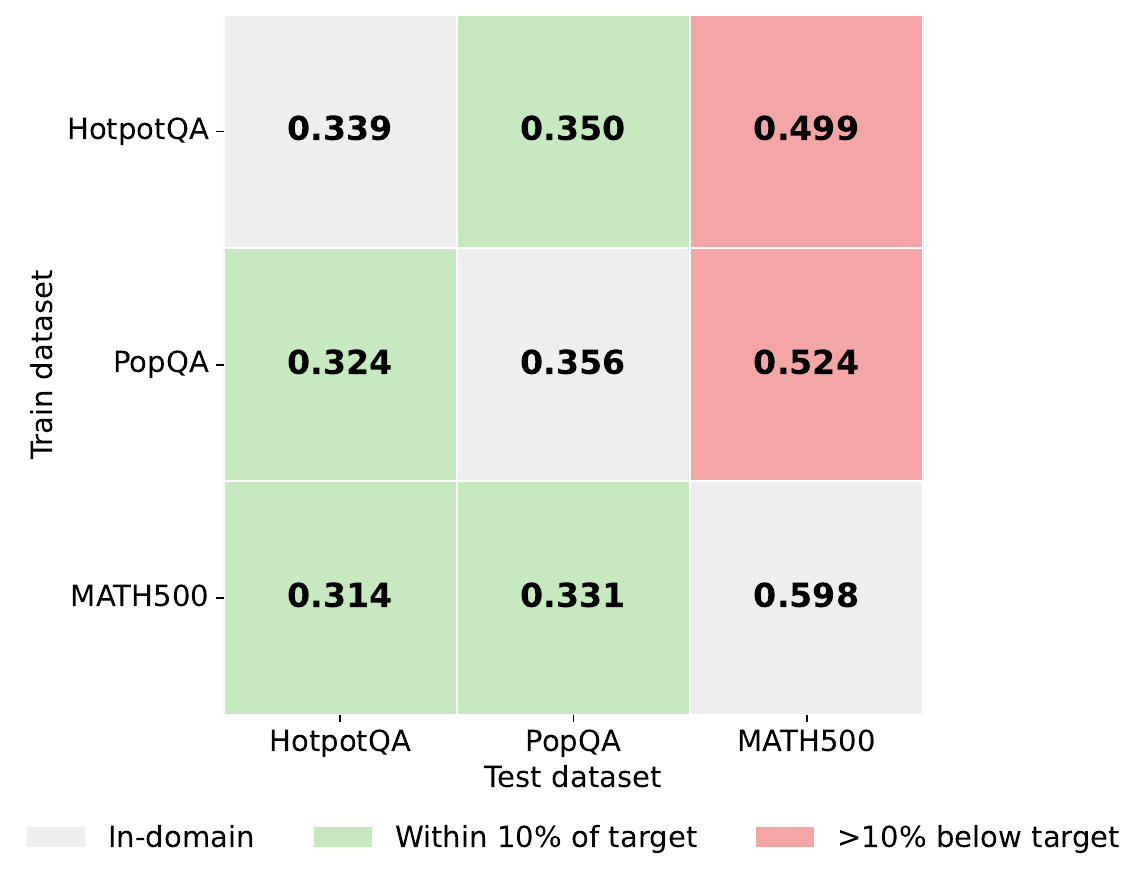}
    \caption{\textbf{Out-of-domain transfer.} Mean ranker quality over 1–99 LLM calls for each train–test dataset pair. Diagonal cells show in-domain performance; off-diagonal cells show out-of-domain transfer. Cells are colored by relative transfer quality: green indicates performance within 10\% of the target dataset’s in-domain score, while red indicates larger degradation. Transfer remains competitive on HotpotQA and PopQA, but drops substantially when transferring to MATH500.}
    \label{fig:transfer_heatmap}
\end{figure}

We evaluate out-of-domain generalization by testing how rankers trained on the training split of one dataset perform on the test split of another dataset (Figure~\ref{fig:transfer_heatmap}). Performance remains competitive on HotpotQA and PopQA, but drops substantially when transferred to MATH500. 

Appendix~\ref{sec:stable_rank_vs_SC} provides some intuition for why several strong baselines perform surprisingly poorly.




\section{Analysis}

\subsection{Feature Ablation}

Figure~\ref{fig:ablation} shows the results of single- and two-feature ablations. On the QA datasets, Answer Length and Answer Centroid Dist are the most individually impactful features, with HotpotQA showing sharper drops than PopQA. On MATH500, no single feature matters much in isolation — the model is robust to any individual removal.

The more interesting pattern emerges from two-feature ablations, which reveal complementarity: removing two features together causes significantly larger drops than the sum of their individual effects. On HotpotQA and PopQA, Ratio to Best and Answer Centroid Dist are the most complementary pair. MATH500 shows the most striking case:  Ratio to Best and Shared Checkpoints are nearly irrelevant individually, yet removing both causes a -5.7\% drop — a strong sign that the ranker relies on their combination rather than either signal alone.



\begin{figure}[t!]
    \centering

    \begin{subfigure}[t]{0.95\linewidth}
        \centering
        \includegraphics[width=\textwidth]{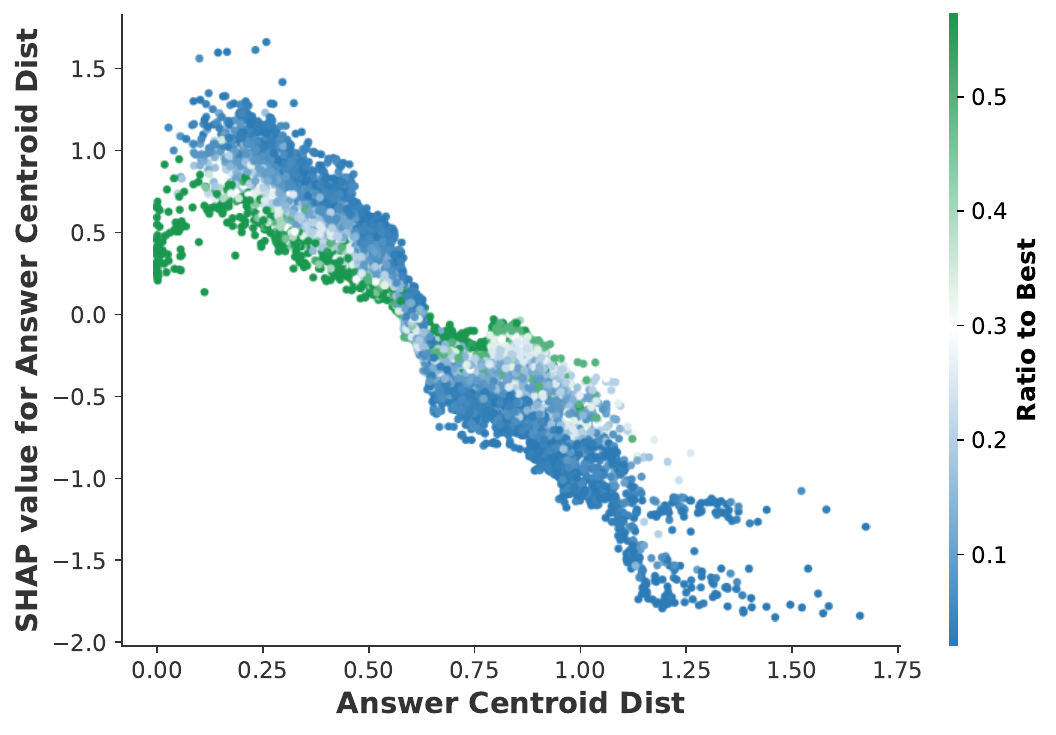}
        \caption{PopQA}
    \end{subfigure}

    \vspace{0.1cm}

    \begin{subfigure}[t]{0.95\linewidth}
        \centering
        \includegraphics[width=\textwidth]{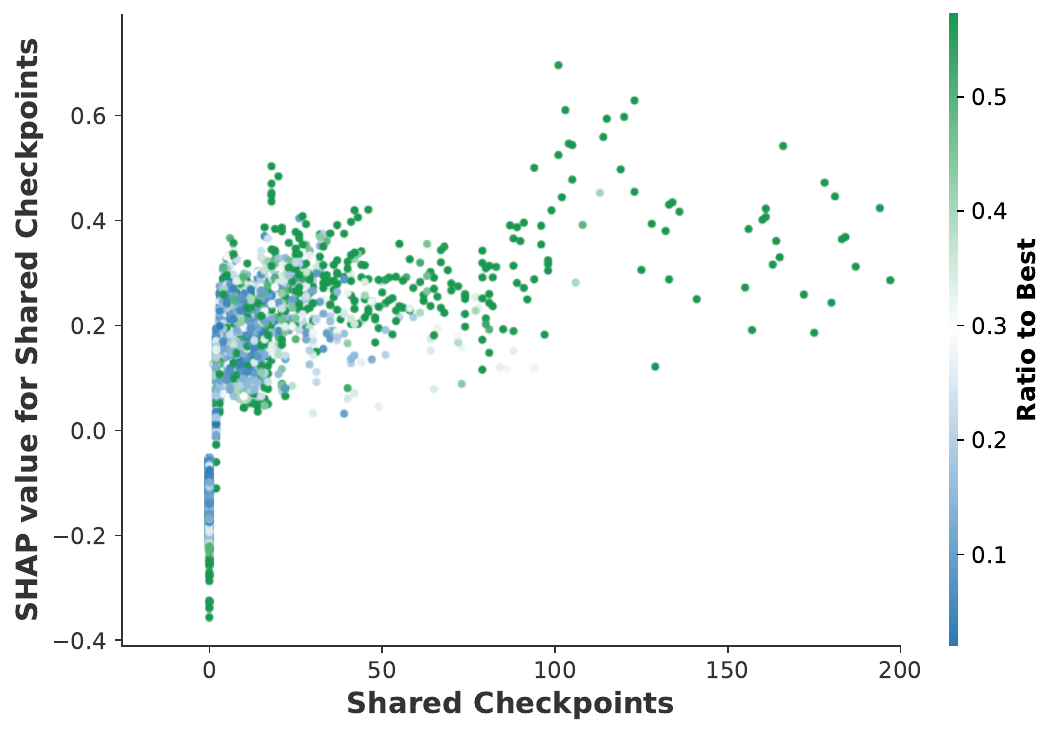}
        \caption{HotpotQA}
    \end{subfigure}

    \vspace{0.1cm}

    \begin{subfigure}[t]{0.95\linewidth}
        \centering
        \includegraphics[width=\textwidth]{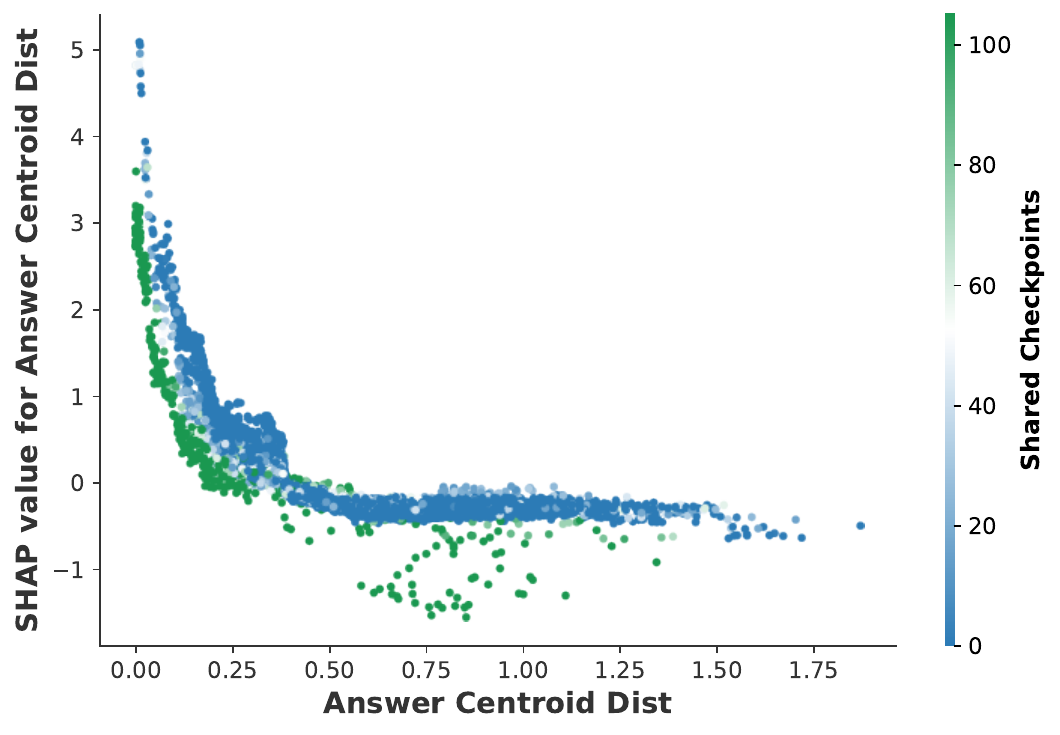}
        \caption{MATH500}
    \end{subfigure}

    \caption{SHAP dependence plots for selected features across PopQA, HotpotQA, and MATH500. The x-axis shows the feature value, the y-axis its contribution to the ranker score, and color encodes a second feature. Vertical color separation at the same x-value reveals that the two features interact — the contribution of one depends on the value of the other.}
    \label{fig:shap_dependence}
\end{figure}

\subsection{SHAP feature analysis}

To further analyze feature interactions, Figure~\ref{fig:shap_dependence} shows SHAP dependence plots for the most influential features per dataset. In each plot, the x-axis is the raw feature value, the y-axis is its SHAP contribution to the ranker score, and the color encodes a second feature. The key signal is vertical color separation: if green and blue points sit at different heights for the same x-value, the two features interact — the effect of the first depends on the value of the second.

The dependence plots reveal that \textit{features do not act independently} — their contributions are conditioned on the values of other features, producing non-additive interactions that vary across datasets. In PopQA, a semantically central answer is rewarded more strongly when it is not the dominant one, while a semantically distant answer is not penalized if it is sufficiently common. In HotpotQA, high  Ratio to Best is penalized when reasoning overlap is weak, suggesting it reflects coincidental rather than genuine agreement. In MATH500, semantically central answers reached via independent reasoning paths are favored over those produced by correlated chains.

\section{Conclusion}

In this study, we presented a new framing of self-consistency as a ranking problem, considering information from the ranking order. Adopting this view, we construct a lightweight ranker trained to rank candidate answers by their correctness. 
Further we test it on three diverse datasets and show that, RISC consistently improves the accuracy-cost trade-off over standard self-consistency and strong baselines. The gains are especially pronounced on the QA datasets, where the ranker frequently matches self-consistency with substantially fewer samples and, at higher budgets, can surpass the best achieved accuracy.

To analyze the ranker’s properties, we conduct a feature analysis showing that the ranker benefits from combining complementary signals rather than relying on any single handcrafted criterion. 
Overall, these results suggest that ranking is a promising and practical direction for improving test-time scaling in LLMs.





\section*{Limitations}

Our study has several limitations. First, although we evaluate the proposed approach using generations from two models, Llama-3.1-8B-Instruct and Olmo-3-7B-Instruct, both are relatively small instruction-tuned models. Therefore, it remains unclear how well the approach generalizes across broader model families and substantially different model scales.

Second, the gains on mathematical reasoning benchmarks are more limited than on question answering tasks. This suggests that the proposed features and ranking formulation may be better suited to settings such as QA, where semantic agreement and answer clustering are more informative, than to domains like math reasoning, where correctness can depend on finer-grained intermediate steps.

Third, although RISC consistently improves over standard self-consistency, there remains a substantial gap to oracle performance. In other words, the correct answer is often present among the sampled candidates, but the ranker still fails to identify it reliably in many cases. This highlights considerable room for improvement in test-time answer selection.

\section*{Acknowledgements}

The work was supported by the grant for research centers in the field of AI provided by the Ministry of Economic Development of the R.F. in accordance with the agreement 000000C313925P4F0002 and the agreement with Skoltech №139-10-2025-033.

\bibliography{custom}

\appendix

\section{Training Details}
\label{sec:train_details}

To improve generalization across inference budgets, we replicate each question across multiple budget settings by retaining only the first \(n\) sampled answers and treating each budget-specific subset as a separate ranking instance, where \(n\) corresponds to the number of available LLM calls.

We exclude training questions that do not induce a meaningful ranking signal, where all candidate answers are either incorrect or correct.
The data is then split into 80\% training and 20\% validation at the original question level, ensuring that budget-specific variants of the same question do not appear in both splits.

After hyperparameter selection on the validation set, we retrain the ranker on the full training data.

Default hyperparameters

\begin{verbatim}
DEFAULT_LGB_PARAMS = dict(
    objective="lambdarank",
    metric="ndcg",
    n_estimators=600,
    learning_rate=0.005,
    max_depth=5,
    num_leaves=20,
    subsample=0.9,
    colsample_bytree=1.0,
    min_data_in_leaf=20,
    feature_fraction_bynode=0.8,
    reg_lambda=1.0,
    reg_alpha=1e-3,
    random_state=42,
)
\end{verbatim}

Hyperparameter grid

\begin{verbatim}
PARAM_GRID = {
    "n_estimators": [500, 600, 700],
    "learning_rate": [0.005, 0.01],
    "num_leaves": [24, 31],
    "max_depth": [5, 6],
    "min_data_in_leaf": [20, 40],
    "reg_lambda": [0.3, 1.0],
    "feature_fraction_bynode": [0.8, 1.0],
}
\end{verbatim}

\section{Baselines vs SC performance}
\label{sec:stable_rank_vs_SC}

\paragraph{Stable Rank vs. Majority Vote.}

Here we provide an additional analysis of why SR preformed poorly. Stable rank captures the geometric spread of a response's hidden-state representation, but this property is only weakly associated with correctness. Wrong (but elaborate) responses can exhibit high representational diversity, while correct (but concise) answers may concentrate on few dimensions.

More fundamentally, stable rank operates as a per-response quality signal and cannot exploit the error-cancellation mechanism that gives majority voting its strength. Correct answers cluster while errors disperse, so as sample size grows, majority voting benefits from stronger consensus. However, SR-based selection becomes increasingly susceptible to outlier responses with anomalously high geometric spread. This explains the observed performance degradation at larger N, and this is why CISC works better. The confidence signal in CISC works because it is fed back into a weighted majority vote, rather than using it for single-response selection. 

\section{SHAP analysis}
\label{sec:shap_analysis}

\begin{figure}[htbp]
    \centering

    \begin{subfigure}[t]{0.5\textwidth}
        \centering
        \includegraphics[width=\textwidth]{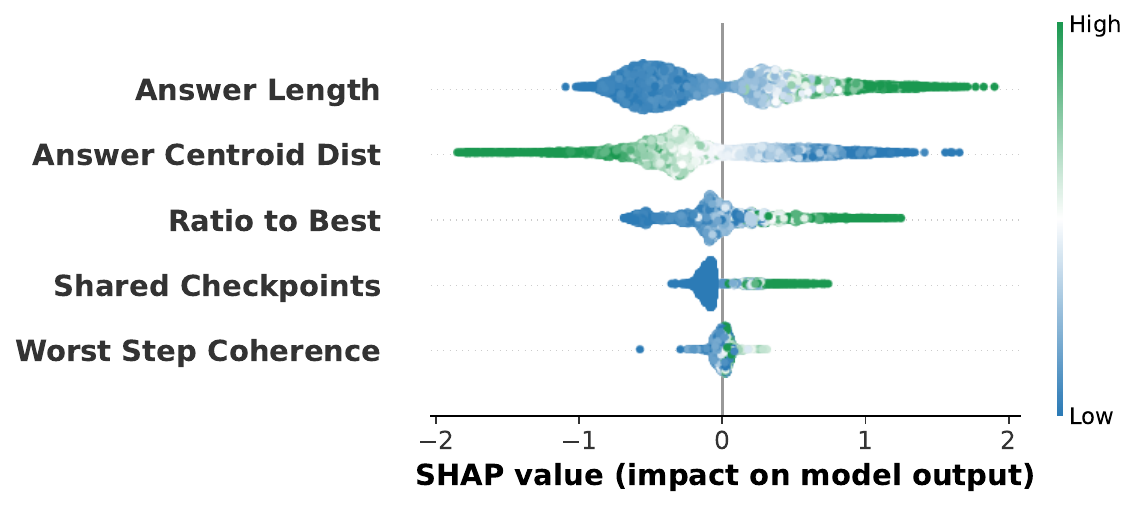}
        \caption{PopQA}
    \end{subfigure}

    \vspace{0.5cm}

    \begin{subfigure}[t]{0.5\textwidth}
        \centering
        \includegraphics[width=\textwidth]{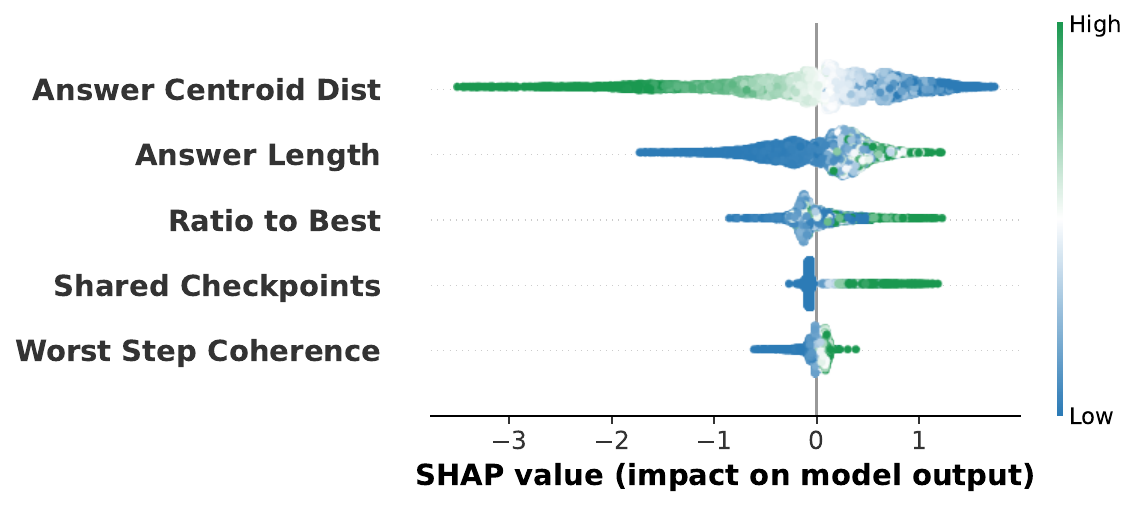}
        \caption{HotpotQA}
    \end{subfigure}

    \vspace{0.5cm}

    \begin{subfigure}[t]{0.5\textwidth}
        \centering
        \includegraphics[width=\textwidth]{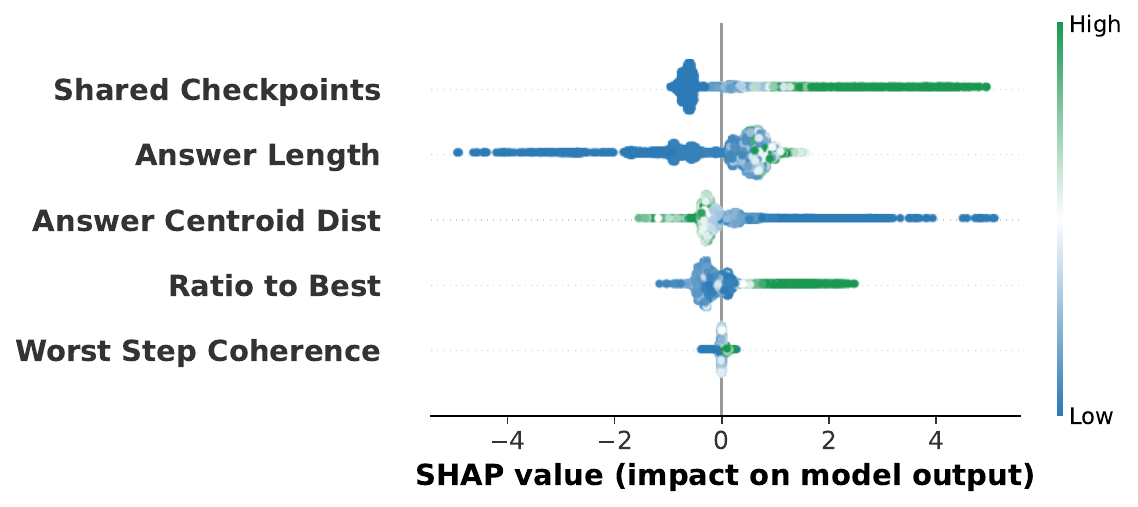}
        \caption{MATH500}
    \end{subfigure}

    \caption{SHAP feature importance plots for three datasets, illustrating the average impact of selected features on the model output across the evaluated samples.}
    \label{fig:shap_FI}
\end{figure}


\newpage
\section{Prompts}
\label{sec:prompts}


\begin{figure}[htbp]
\begin{tcolorbox}[
    breakable,
    colback=gray!6,
    colframe=black!40,
    arc=3pt,
    boxrule=0.5pt,
    title={\small \textbf{Self-Consistency Prompt Template (PopQA)}},
    fonttitle=\bfseries
]

\textbf{\small [System]} \\
\small You are a helpful reasoning assistant for question answering. 
Answer the user's question directly based on your knowledge.

\tcblower

\textbf{\small [User]} \\
\small Reason step by step very concisely before giving a final answer. \\
Strictly follow the format: \\[2pt]
\texttt{\small Step-by-step reasoning: [your reasoning]} \\
\texttt{\small Final Answer: [your answer]} \\[6pt]
Question: \textit{\{question\}}

\end{tcolorbox}
\caption{Prompt template used for self-consistency chain-of-thought sampling. 
The model is queried $N$ times with temperature $T > 0$; 
the final answer is then selected by majority vote over extracted answers.}
\label{fig:prompt_sc}
\end{figure}

\begin{figure}[htbp]
\begin{tcolorbox}[
    breakable,
    colback=gray!6,
    colframe=black!40,
    arc=3pt,
    boxrule=0.5pt,
    title={\small \textbf{MATH500 Prompt Template}},
    fonttitle=\bfseries
]

\textbf{\small [System]} \\
\small You are a helpful and expert mathematical assistant. 
Solve the given math problem step-by-step.

\tcblower

\textbf{\small [User]} \\
\small Reason step-by-step to solve the math problem. \\
At the end of your solution, strictly follow this format to present your final answer: \\[2pt]
\texttt{\small Final Answer: \textbackslash boxed\{answer\}} \\[2pt]
\small Ensure the answer within \textbackslash boxed\{\} is concise and contains only the final value or expression. \\[6pt]
Problem: \textit{\{problem\}}

\end{tcolorbox}
\caption{Prompt template used for the MATH500 dataset. 
The system prompt establishes the role of a mathematical assistant, 
while the user instructions enforce a chain-of-thought structure with a specific \texttt{\textbackslash boxed\{\}} format for the final answer extraction.}
\label{fig:prompt_math500}
\end{figure}

\begin{figure}[htbp]
\begin{tcolorbox}[
    colback=gray!6,
    colframe=black!40,
    arc=3pt,
    boxrule=0.5pt,
    title={\small \textbf{Self-Consistency Prompt Template (HotpotQA)}},
    fonttitle=\bfseries
]

\textbf{\small [System]} \\
\small You are a helpful assistant that answers questions by reasoning step by step.

\tcblower

\textbf{\small [User]} \\
\small Think through the question carefully, then provide your final answer \\[2pt]
\small as a short phrase on the last line in the format: \\
\texttt{\small Final Answer: [your answer]} \\[6pt]
Question: \textit{\{question\}}

\end{tcolorbox}
\caption{Prompt template used for self-consistency chain-of-thought sampling. 
The model is queried $N$ times with temperature $T > 0$; 
the final answer is then selected by majority vote over extracted answers.}
\label{fig:prompt_sc_hotpot}
\end{figure}




\section{ReASC algorithm}
\label{appendix:reasc}

Given an input $x$, ReASC~\cite{reasc} samples responses $y_i \sim p_\theta(\cdot \mid x)$ with extracted answers $a_i=\mathrm{Ans}(y_i)$ and confidence scores $s_i=S(y_i)$. If the first sampled response satisfies $s_1 \ge \tau_{\mathrm{gate}}$, the algorithm returns $a_1$. Otherwise, it accumulates weighted evidence
\[
v(a)=\sum_{i:\,a_i=a}\max\!\left(1,\exp\!\left(\lambda \frac{s_i-\mu}{\sigma}\right)\right),
\]
where $(\mu,\sigma)$ are calibration statistics and $\lambda$ controls sensitivity to confidence. After each sample, letting $a^{(1)}$ and $a^{(2)}$ denote the answers with the largest and second-largest evidence, ReASC defines
\[
\alpha=v(a^{(1)})+1,\qquad \beta=v(a^{(2)})+1,
\]
and stops when
\[
1-I_{1/2}(\alpha,\beta)\ge C_{\mathrm{thresh}}.
\]
The final prediction is $\hat a=\arg\max_a v(a)$.

\onecolumn

\section{Example of Worst-Step Coherence feature}

Here, we present a comparison between two traces for the same question–answer pair. However, in the actual calculations, these computations are performed over all traces for each question–answer pair, rather than just the two traces shown here.

\label{appendix:worst_step}

\vspace{1.5ex}

\textbf{Question:} The football manager who recruited David Beckham managed Manchester United during what timeframe?

\subsection*{Good Chain}

\goodstepup{0.890848}{65}{To answer this question, I need to identify the football manager who recruited David Beckham and then determine during which timeframe he managed Manchester United.}

\goodstepup{0.760034}{29}{First, I know that David Beckham is a retired English football player who played for several clubs, including Manchester United.}

\goodloweststepup{0.740996}{24}{He was recruited to Manchester United in 1992, when he was just 17 years old.}

\goodstepup{0.826989}{47}{Next, I need to identify the manager who signed David Beckham to Manchester United.}

\goodstepup{0.869573}{59}{After some research, I found that the manager who recruited David Beckham was Alex Ferguson.}

\goodstepup{0.807464}{42}{Alex Ferguson is a well-known former manager of Manchester United, and he played a significant role in shaping the club's history during his 26-year tenure.}

\goodstepup{0.885220}{63}{Finally, to determine the timeframe during which Alex Ferguson managed Manchester United, I need to consider that David Beckham arrived at the club in 1992.}

\goodstepup{0.798782}{40}{Alex Ferguson's tenure as manager of Manchester United began in 1986, and he remained in the position until his retirement in 2013.}

\goodstepup{0.839749}{51}{Therefore, considering all the information, I can conclude that Alex Ferguson managed Manchester United during the timeframe from 1986 to 2013.}

\vspace{2ex}

\finalanswer{The answer is: from 1986 to 2013.}

\subsection*{Bad Chain}

\badstepup{0.816500}{44}{To determine the timeframe during which the football manager who recruited David Beckham managed Manchester United, let's consider the key information.}

\badstepup{0.821974}{46}{David Beckham played for Manchester United from 1992 to 2003.}

\badloweststepup{0.727240}{20}{Given this, it is likely that the manager we are looking for is Sir Alex Ferguson.}

\badstepup{0.927873}{75}{Sir Alex Ferguson was the manager at Manchester United during the entire period in which David Beckham played for the team.}

\badstepup{0.905895}{69}{Sir Alex Ferguson is widely known to have recruited David Beckham during his tenure, when Beckham moved from Manchester United's youth team to the first team in 1992.}
\vspace{2ex}

\finalanswer{The answer is: Sir Alex Ferguson.}

\section{Example of Shared Checkpoints feature}

Here, we present a comparison between two traces for the same question–answer pair. However, in the actual calculations, these computations are performed over all traces for each question–answer pair, rather than just the two traces shown here.

\label{appendix:shared_checkpoints}

\begin{tcolorbox}[breakable,title={Trace A vs Trace B for the math problem: color-coded shared checkpoints}]

\textbf{Problem.}
Let
\[
x^8 + 3x^4 - 4 = p_1(x)p_2(x)\dotsm p_k(x),
\]
where each non-constant polynomial \(p_i(x)\) is monic with integer coefficients and cannot be factored further over the integers. Compute
\[
p_1(1) + p_2(1) + \dots + p_k(1).
\]

\medskip
\hrule
\medskip
\textbf{Trace A}\\[0.5em]
{\sethlcolor{pairfour}\hl{Step 1:  The problem requires us to factor the given polynomial \$x\textasciicircum{}8 + 3x\textasciicircum{}4 - 4\$ into irreducible polynomials and then find the sum of the values obtained by evaluating each of these polynomials at \$x = 1\$.}}\textsuperscript{4}\\ 
Step 2:  To start, we can try to factor the given polynomial using the sum of squares identity, \$a\textasciicircum{}2 - b\textasciicircum{}2 = (a - b)(a + b)\$. We notice that \$x\textasciicircum{}8 + 3x\textasciicircum{}4 - 4\$ can be expressed as \$(x\textasciicircum{}4)\textasciicircum{}2 + 2x\textasciicircum{}4 + 1 - 2x\textasciicircum{}4 - 4\$, which simplifies to \$(x\textasciicircum{}4 + 1)\textasciicircum{}2 - (x\textasciicircum{}2)\textasciicircum{}2\$. Applying the sum of squares identity, we get \$(x\textasciicircum{}4 + 1 - x\textasciicircum{}2)(x\textasciicircum{}4 + 1 + x\textasciicircum{}2)\$.\\ 
Step 3:  Further simplifying, we employ the difference of squares again: \$x\textasciicircum{}4 + 1 - x\textasciicircum{}2 = (x\textasciicircum{}2 - x + 1)(x\textasciicircum{}2 + x + 1)\$, and \$x\textasciicircum{}4 + 1 + x\textasciicircum{}2 = (x\textasciicircum{}2 + x + 1)(x\textasciicircum{}2 - x + 1)\$.\\ 
{\sethlcolor{pairone}\hl{Step 4:  We see that the factor \$(x\textasciicircum{}2 + x + 1)\$ appears in both factors. Factoring this out first allows us to obtain the factorization: \$(x\textasciicircum{}4 + 3x\textasciicircum{}2 - 4) = (x\textasciicircum{}2 + x + 1)(x\textasciicircum{}2 - x + 1)(x\textasciicircum{}2 + 4)\$.}}\textsuperscript{1}\\ 
{\sethlcolor{pairtwo}\hl{Step 5:  Now, we need to evaluate each of these factors at \$x = 1\$. When we substitute \$x = 1\$ into each factor, we get \$1 + 1 + 1 = 3\$ from the first polynomial, \$1 - 1 + 1 = 1\$ from the second polynomial, and \$1 + 4 = 5\$ from the last polynomial.}}\textsuperscript{2}\\ 
{\sethlcolor{pairthree}\hl{Step 6:  The sum of these values is \$3 + 1 + 5 = \textbackslash\{\}boxed\{9\}\$.}}\textsuperscript{3}

\tcblower

\textbf{Trace B}\\[0.5em]
\#\# {\sethlcolor{pairfour}\hl{Step 1:  The given expression is \$x\textasciicircum{}8 + 3x\textasciicircum{}4 - 4\$. To factorize this expression into the product of non-constant polynomials \$p\_i(x)\$, we can use the difference of squares factorization.\\ 
\#\#}}\textsuperscript{4} Step 2:  We can rewrite \$x\textasciicircum{}8 + 3x\textasciicircum{}4 - 4\$ as a difference of squares: \$(x\textasciicircum{}4 + 4)(x\textasciicircum{}4 - 1)\$, where the latter expression can be further factorized as a difference of squares.\\ 
\#\# Step 3:  Now, we apply the difference of squares identity, which gives us: \$(x\textasciicircum{}4 - 1) = (x\textasciicircum{}2 + 1)(x + 1)(x - 1)\$.\\ 
\#\# Step 4:  Combining the initial factorization with the further factorization, we get the complete factorization of \$x\textasciicircum{}8 + 3x\textasciicircum{}4 - 4 = (x\textasciicircum{}4 + 4)(x\textasciicircum{}2 + 1)(x + 1)(x - 1)\$.\\ 
\#\# {\sethlcolor{pairone}\hl{Step 5:  Finally, since \$x\textasciicircum{}4 + 4\$ cannot be factored further over the integers, we have \$k = 4\$, and the polynomials are \$p\_1(x) = x\textasciicircum{}4 + 4\$, \$p\_2(x) = x\textasciicircum{}2 + 1\$, \$p\_3(x) = x + 1\$, and \$p\_4(x) = x - 1\$.\\ 
\#\#}}\textsuperscript{1} {\sethlcolor{pairtwo}\hl{Step 6:  We can now evaluate each \$p\_i(1)\$ by substituting \$x = 1\$ into each of the polynomials. This gives us \$p\_1(1) = (1)\textasciicircum{}4 + 4 = 5\$, \$p\_2(1) = (1)\textasciicircum{}2 + 1 = 2\$, \$p\_3(1) = 1 + 1 = 2\$, and \$p\_4(1) = 1 - 1 = 0\$.\\ 
\#\#}}\textsuperscript{2} {\sethlcolor{pairthree}\hl{Step 7:  Therefore, we find that \$p\_1(1) + p\_2(1) + p\_3(1) + p\_4(1) = 5 + 2 + 2 + 0 = 9\$.}}\textsuperscript{3}
\end{tcolorbox}

\begin{longtable}{>{\raggedright\arraybackslash}p{0.06\textwidth}
                  >{\centering\arraybackslash}p{0.08\textwidth}
                  >{\centering\arraybackslash}p{0.10\textwidth}
                  >{\raggedright\arraybackslash}p{0.35\textwidth}
                  >{\raggedright\arraybackslash}p{0.35\textwidth}}

\hline
\textbf{Pair} & \textbf{Color} & \textbf{Similarity} & \textbf{Checkpoint in Trace A} & \textbf{Checkpoint in Trace B} \\
\hline
\endfirsthead

\hline
\textbf{Pair} & \textbf{Color} & \textbf{Similarity} & \textbf{Checkpoint in Trace A} & \textbf{Checkpoint in Trace B} \\
\hline
\endhead

1 & \fcolorbox{black}{pairone}{\makebox[1.5em]{\strut}}  & 0.823 & Step 4:  We see that the factor \$(x\textasciicircum{}2 + x + 1)\$ appears in both factors. Factoring this out first allows us to obtain the factorization: \$(x\textasciicircum{}4 + 3x\textasciicircum{}2 - 4) = (x\textasciicircum{}2 + x + 1)(x\textasciicircum{}2 - x + 1)(x\textasciicircum{}2 + 4)\$. & Step 5:  Finally, since \$x\textasciicircum{}4 + 4\$ cannot be factored further over the integers, we have \$k = 4\$, and the polynomials are \$p\_1(x) = x\textasciicircum{}4 + 4\$, \$p\_2(x) = x\textasciicircum{}2 + 1\$, \$p\_3(x) = x + 1\$, and \$p\_4(x) = x - 1\$.
\#\# \\
2 & \fcolorbox{black}{pairtwo}{\makebox[1.5em]{\strut}}  & 0.823 & Step 5:  Now, we need to evaluate each of these factors at \$x = 1\$. When we substitute \$x = 1\$ into each factor, we get \$1 + 1 + 1 = 3\$ from the first polynomial, \$1 - 1 + 1 = 1\$ from the second polynomial, and \$1 + 4 = 5\$ from the last polynomial. & Step 6:  We can now evaluate each \$p\_i(1)\$ by substituting \$x = 1\$ into each of the polynomials. This gives us \$p\_1(1) = (1)\textasciicircum{}4 + 4 = 5\$, \$p\_2(1) = (1)\textasciicircum{}2 + 1 = 2\$, \$p\_3(1) = 1 + 1 = 2\$, and \$p\_4(1) = 1 - 1 = 0\$.
\#\# \\
3 & \fcolorbox{black}{pairthree}{\makebox[1.5em]{\strut}}  & 0.823 & Step 6:  The sum of these values is \$3 + 1 + 5 = \textbackslash\{\}boxed\{9\}\$. & Step 7:  Therefore, we find that \$p\_1(1) + p\_2(1) + p\_3(1) + p\_4(1) = 5 + 2 + 2 + 0 = 9\$. \\
4 & \fcolorbox{black}{pairfour}{\makebox[1.5em]{\strut}}  & 0.800 & Step 1:  The problem requires us to factor the given polynomial \$x\textasciicircum{}8 + 3x\textasciicircum{}4 - 4\$ into irreducible polynomials and then find the sum of the values obtained by evaluating each of these polynomials at \$x = 1\$. & Step 1:  The given expression is \$x\textasciicircum{}8 + 3x\textasciicircum{}4 - 4\$. To factorize this expression into the product of non-constant polynomials \$p\_i(x)\$, we can use the difference of squares factorization.
\#\# \\

\hline
\end{longtable}

\section{Inference on the Olmo-3-7B-Instruct model}

\label{inference_olmo}
\begin{figure*}[htb!]
    \centering
    \includegraphics[trim=.25cm .35cm .5cm .25cm, clip, width=\linewidth]{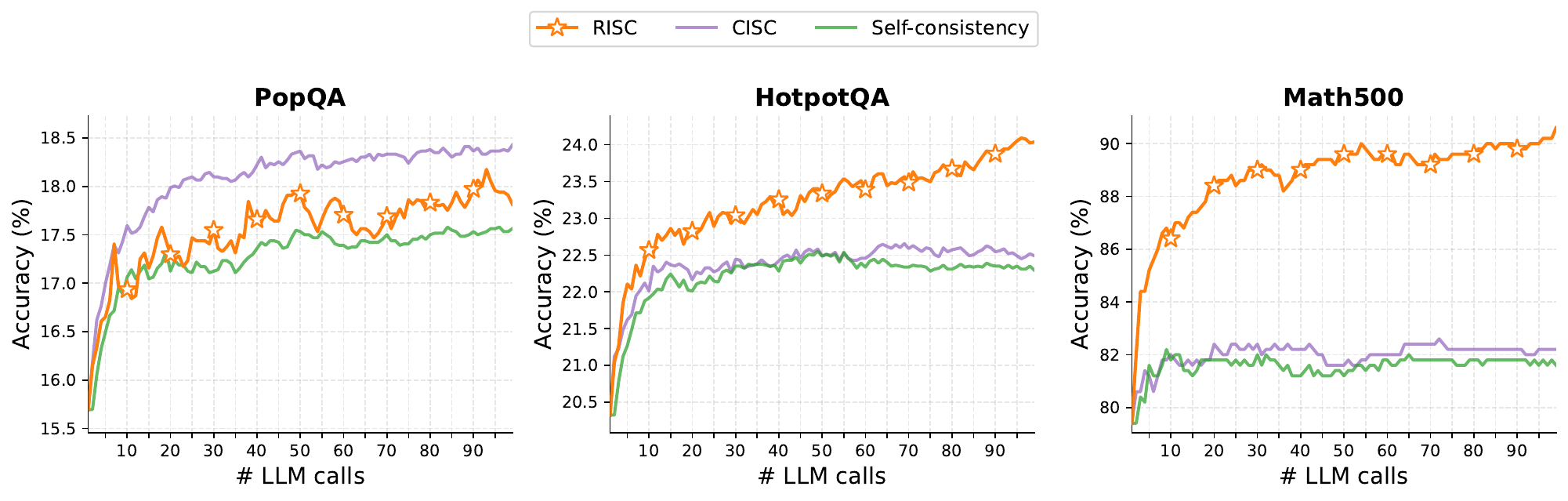}
    \caption{Comparison of RISC against Self-Consistency, Stable Rank, ReASC, and CISC on three datasets for Olmo-3-7B-Instruct model. RISC consistently outperforms the baselines on the HotpotQA and MATH500 datasets.
    }
    \label{fig:baselines_comparison_olmo}
    \vspace{-0.5cm}
\end{figure*}

\vspace{2ex}

Compared to the results obtained with Llama-3.1-8B-Instruct, we observe a performance drop on PopQA, which is likely due to the disproportionately small number of high-quality training examples, i.e., questions for which both correct and incorrect candidate answers are available. For HotpotQA, we observe a similar trend. In contrast, on MATH500, the method shows a substantial performance improvement compared to Llama-3.1-8B-Instruct.  The majority answer is often already correct, as reflected by the top-1 accuracy. However, among the remaining cases, errors frequently occur when the majority answer is wrong because generations stop at an intermediate reasoning step without producing the final answer. Since p-true-based CISC may still assign high confidence to such incomplete reasoning traces, RISC handles these cases more effectively. Since such cases are the main source of top-1 errors when the majority choice is wrong, correcting them yields a large apparent performance gain.

\end{document}